\documentclass[10pt, a4paper]{article}

\usepackage{lrec-coling2024} 

\usepackage{booktabs}
\usepackage{algpseudocode}
\usepackage{algorithm}
\usepackage{listings}
\usepackage{amsfonts}
\usepackage{mathtools}
\usepackage{amsmath}
\usepackage{todonotes}
\usepackage{float}
\usepackage{longtable}
\usepackage{setspace}
\usepackage{multicol}
\usepackage{amsbsy}
\usepackage{caption}
\usepackage{subcaption}
\usepackage[capitalise, noabbrev]{cleveref}
\usepackage[shortcuts, acronym, hyperfirst = false]{glossaries}
\glsdisablehyper

\title{ 
\textbf{Intrinsic Subgraph Generation for Interpretable Graph based Visual Question Answering}
}

\name{Pascal Tilli, Ngoc Thang Vu} 

\address{
    Institute for Natural Language Processing (IMS) \\
    University of Stuttgart \\
    Stuttgart, Germany \\
    \{pascal.tilli, thang.vu\}@ims.uni-stuttgart.de\\
}

\definecolor{Newcolor}{rgb}{0, 0.8, 0.6}
\definecolor{Blackcolor}{rgb}{0, 0, 0}

\begin{document}
\newacronym{gnn}{GNN}{Graph Neural Network}
\newacronym{gnns}{GNNs}{Graph Neural Networks}
\newacronym{sgg}{SGG}{Scene Graph Generation}
\newacronym{vqa}{VQA}{Visual Question Answering}
\newacronym{qa}{QA}{Question Answering}
\newacronym{dl}{DL}{Deep Learning}
\newacronym{ml}{ML}{Machine Learning}
\newacronym{nlp}{NLP}{Natural Language Processing}
\newacronym{cv}{CV}{Computer Vision}
\newacronym{gat}{GAT}{Graph Attention Network}
\newacronym{gats}{GATs}{Graph Attention Networks}
\newacronym{xai}{XAI}{Explainable Artificial Intelligence}
\newacronym{imle}{I-MLE}{Implicit Maximum Likelihood Estimation}
\newacronym{nsm}{NSM}{Neural State Machine}
\newacronym{rl}{RL}{Reinforcement Learning}
\newacronym{mgat}{M-GAT}{Masking Graph Attention Network}
\newacronym{subgat}{Sub-GAT}{Subgraph Graph Attention Network}
\newacronym{kergnns}{KerGNNs}{Kernel Graph Neural Networks}
\newacronym{wl}{WL}{Weisfeiler-Lehman}
\newacronym{cnn}{CNN}{Convolutional Neural Network}
\newacronym{cnns}{CNNs}{Convolutional Neural Networks}
\newacronym{mlp}{MLP}{Multi-Layer Perceptron}
\newacronym{gnnex}{GNNExplainer}{GNNExplainer}
\newacronym{pyg}{PyG}{PyTorch Geometric}
\newacronym{atcoo}{AT-COO}{Answer Token Co-occurrence}
\newacronym{qtcoo}{QT-COO}{Question Token Co-occurrence}

\abstract{
The large success of deep learning based methods in Visual Question Answering (VQA) has concurrently increased the demand for explainable methods.
Most methods in Explainable Artificial Intelligence (XAI) focus on generating post-hoc explanations rather than taking an intrinsic approach, the latter characterizing an interpretable model.
In this work, we introduce an interpretable approach for graph-based VQA and demonstrate competitive performance on the GQA dataset.
This approach bridges the gap between interpretability and performance.
Our model is designed to intrinsically produce a subgraph during the question-answering process as its explanation, providing insight into the decision making.
To evaluate the quality of these generated subgraphs, we compare them against established post-hoc explainability methods for graph neural networks, and perform a human evaluation.
Moreover, we present quantitative metrics that correlate with the evaluations of human assessors, acting as automatic metrics for the generated explanatory subgraphs.
Our implementation is available at \url{https://github.com/DigitalPhonetics/Intrinsic-Subgraph-Generation-for-VQA}.
 \\ \newline 
}

\maketitleabstract

\section{Introduction}
\label{sec:intro}

\gls{vqa} \cite{vqa, vqa_2} is acknowledged as a challenging multi-modal task for \gls{ml} algorithms as it requires a semantic comprehension of images in relation to the posed questions.
State-of-the-art approaches for \gls{vqa} are systems based on \gls{dl}, which are mostly assessed by accuracy and efficiency metrics.
To enhance collaboration between humans and \gls{ml} systems in real-world settings, it is essential to reliably comprehend the system's outputs.
Nonetheless, the majority of \gls{dl} based \gls{vqa} systems are considered black boxes by both users and developers.
Hence, deploying these systems in important decision-making domains carries considerable risk, despite their progress.

The domain of \gls{xai} addresses the aforementioned issue, and can be categorized into two subdomains: \emph{explainability} and \emph{interpretability} \cite{rudin2019stop, marcinkevivcs2020interpretability}.
The domain of interpretability centers on inherently interpretable models, i.e. where the decision making process of the models can be comprehended by humans, e.g. decision trees.
Explainable \gls{ml} focuses on methods that generate explanations post-hoc for already existing (black-box) models,
possibly requiring additional hyperparameter tuning.
Notably, interpretable models possess the advantage of intrinsic explanation generation, where the model itself generates explanations. 
This contrasts with post-hoc methods, which introduce an additional method or model aimed at explaining predictions, leading to increased computational cost.

Explainability methods for \gls{vqa} often focus on pixel importance as visual explanations \cite{clevr_xai, vqa_grad_cam}.
Some approaches address explainability by generating rationales to explain the system's predicted answers.
These rationales can be generated either post-hoc by a another neural network or by the original network itself \cite{aokvqa}.
However, it remains uncertain whether the answer and rationale generation processes influence one another.
An alternative strategy involves formulating contrastive explanations \cite{clevr_xai}.
Moreover, some models that yield intermediate outputs are considered to offer interpretability \cite{survey_vqa_xai}.
For instance, in \cite{inter_vqa}, the system translates images into textual descriptions relevant to the given question, uses these to predict the answer and is thereby interpretable.
These methods purport to offer interpretability, yet they do not align with our definition wherein an interpretable model should intrinsically generate its own explanation.

In this work, we introduce an approach for graph based \gls{vqa}  that employs a structured graph representation of the displayed scene instead of the raw image input.
The primary goal of our work, is to generate a subgraph alongside the model's prediction as explanation, highlighting the most relevant nodes for a given question, as opposed to employing post-hoc explanation methods.

We focus on the following research questions:
\begin{description}
    \item[\textbf{RQ1}] How can we increase the interpretability of deep learning-based VQA answer prediction through the utilization of Graph Neural Networks?
    \item[\textbf{RQ2}] How does the quality of explanations generated by our method compare to that of state-of-the-art post-hoc explanation methods when evaluated by human assessors?
    \item[\textbf{RQ3}] What methods can we employ to quantitatively assess the quality of explanations in cases where no ground-truth references are available, and to what extent do these quantitative measures align with human preferences?
\end{description}

To address these research questions, we propose a system featuring a \gls{gat} at its core component, which is able to extract a subgraph as explanation for the prediction.
We validate our approach in a graph based \gls{vqa} setting using GQA \cite{gqa_2}.
Furthermore, we conduct a human evaluation to compare our internally generated subgraph with explanations generated by post-hoc methods.
Additionally, we introduce evaluation metrics tailored to subgraphs used as explanations in scenarios where ground-truth explanations are unavailable.

Our contributions can be summarized as follows:
\begin{enumerate}
    \item We propose a novel VQA system that not only provides answers but also offers relevant explanations. Our approach has been proven to deliver highly accurate results, and human evaluators have shown a preference for our intrinsic explanations over traditional post-hoc explainability methods.
    \item We introduce quantitative metrics, which correlate with the results of the human evaluation, to measure the quality of explanations.
\end{enumerate}

\section{Related Work}
\label{sec:related_work}
\subsection{Graph Neural Networks}
\gls{gnns} are evolving to an increasingly more popular area of research due to their recent successes \cite{gnn_survey_2, gnn_survey_3}.
They are designed to harness graph-structured data and perform message passing among nodes to learn contextualized node embeddings.
In addition to their natural applicability in domains where it is obvious to represent data as graphs, such as chemistry (where molecules can be modeled as graphs) or e-commerce (where users interactions can be represented as a graph) \cite{gnn_survey_3}, \gls{gnns} have been successfully applied in \gls{nlp} tasks \cite{gnn_nlp_survey} and \gls{cv}, including tasks like \gls{sgg} \cite{gnn_survey_3}.
For a comprehensive overview spanning various domains, readers are directed  to \citet{gnn_survey_3} regarding a general survey on \gls{gnns}.
For in-depth exploration with a focus on \gls{nlp} \citet{gnn_nlp_survey} offer an extensive resource.

\subsection{Visual Question Answering}
Many different datasets \cite{clevr, gqa_2, vqa_cp, text_vqa} and models have been published since the task itself has been introduced, and aim to evaluate divers capabilities of models \cite{vqa_benchmarking, vqa_medical, shao2023prompting}.
Given the graph-based nature of our \gls{vqa} approach, the subsequent section will be dedicated exclusively to this aspect.

\subsection{Graph based VQA Models}
Graph based \gls{vqa} represents a specialized variant of \gls{vqa} where models use intermediate graph structures that represent the scenes in images, which enables the usage of powerful \gls{gnns} for the task.
\citet{nsm} proposed a \gls{nsm}, which incorporated question guided traversal of scene graphs.
This approach treats nodes as states and allows transitions along edges (relations) in the graph.
Subsequently, \citet{graph_vqa} extended this notion by employing instruction vectors to guide the information propagation of a \gls{gnn}.
The authors conducted comparisons utilizing ground-truth scene graphs sourced from the GQA dataset \cite{gqa_2}.
Remarkably, the \gls{gat} \cite{gat, gat_v2} achieved the best performance by a substantial margin.
In the work of \citet{graphhopper} a \gls{rl} based approach was developed, treating \gls{vqa} as a graph traversal problem.
Their reported performance on ground-truth scene graphs within the GQA dataset slightly trailed behind the \gls{gat} approach by \citet{graph_vqa}.

\citet{vqa_gnn_sgg_1} proposed an approach that centers on a \gls{sgg} model, which transforms the scene graph into two graphs.
One of these graphs emphasizes objects, while the other focuses on the relational aspects. 
\citet{vqa_gnn_sgg_2} is characterized by a bidirectional fusion process between unstructured and structured multimodal knowledge to obtain unified knowledge representation. This fusion ultimately yields a unified knowledge representation.
Both have a different focus compared to our approach prioritizing the \gls{sgg} task, particularly in terms of the methods applied for crafting the scene graph.

The interpretability or explainability of such systems is often asserted without being tested, i.e. the human for whom the explanation is intended is left out of the evaluation loop.
\citet{zhu2022shallow} construct three types of graphs during the reasoning process to generate an answer.
Although the model is described as interpretable because humans can view the intermediate graph representations, there is no actual assessment of its interpretability conducted by human evaluators.
Similarly, \citet{sarkisyan2022graph} assert that they construct an interpretable model by mapping questions into a graph structures which is supposed to make it interpretable.
However, no tests are carried out to assess this aspect of interpretability.

\citet{zeroshot_masks} introduces an approach for knowledge-based \gls{vqa}, where masking strategies are applied to predict answers to better generalize to out-of-distribution answers.
In contrast, our approach employs a differentiable hard-attention masking to extract a subgraph during the message-propagation of the \gls{gnn}.

\subsection{Explainability and Interpretability}
Interpretability and explainability have garnered increased attention alongside the successes of \gls{ml} methods across diverse domains.
\citet{xai_survey_1} conducted a review focusing on \gls{xai} and interpretable \gls{ml} methods.
In addition to their comprehensive literature review, they introduced a systematic taxonomy of these approaches, coupled with references to their programming implementations.

\citet{interpretability_gnns} reviewed the interpretability in \gls{gnns}.
Notably, most approaches for \gls{gnns} are post-hoc methods that try to explain predictions of black-box models.
Transparent approaches in this context are relatively rare.
\gls{dl} methods are recognized as interpretable or transparent that predominantly rely on soft attention scores.
Efforts have been dedicated to investigate if soft attention can be considered interpretable, yielding findings that challenge their alignment with true interpretability \cite{serrano2019attention}.

\citet{kernel_gnns} proposed \gls{kergnns} that integrate kernels into the message passing process of \gls{gnns} in order to overcome the limitation of standard \gls{gnns} which are not capable of surpassing the performance of the \gls{wl} algorithm in a graph isomorphism test.

Drawing inspiration from the idea of visualizing filters of \gls{cnns} \cite{cnn} to identify important feature regions, they visualize the trained kernel based filters to detect key structures in the input and thus referring to their method being interpretable.

\subsection{Soft- and Hard-Attention}
Soft-attention \cite{soft_attention} is generally defined by a continuous variable, while hard-attention is defined by a discrete variable.
Consequently, we can employ gradient-descent methods to differentiate soft-attention, but we cannot utilize these methods for hard-attention due to the non-differentiable nature of a discrete step.
In contrast, hard-attention operates with discrete values, which is typically implemented using sampling methods such as the Gumbel-Softmax \cite{gumbel_softmax_1, gumbel_softmax_2} or \gls{rl} techniques.

In terms of interpretability, hard attention's binary nature makes results clearer and reduces ambiguity about important data parts.

\citet{imle} proposed a framework for backpropagating through such a discrete sampling step.
Their method can be described as a general-purpose algorithm for hybrid learning of systems that contain discrete components embedded in a computational graph.
In our specific application, we leverage \gls{imle} to approximate the gradients for a discrete \emph{top-k} sampling procedure of nodes in a scene graph.

\section{Problem Formulation}
Our problem is framed as a graph representation learning challenge within the multi-modal domain of \gls{vqa}.
Typically, \gls{vqa} is characterized by the function $f(q, i) \rightarrow a$, where $q$ represents a question, $i$ corresponds to an image, and $a$ signifies the answer (treated as a classification problem, with $a \in \mathbb{R}^{n_a}$, where $n_a$ denotes the predefined number of potential answers).

In our case, we depart from the conventional image $i$ and instead employ a graph $g$ as the visual representation, characterized as: $f(q, g) \rightarrow a$.
The question is represented as a sequence of tokens $q \in \mathbb{R}^{n_q \times d_q}$, where $n_q$ signifies the number of tokens and $d_q$ the dimensionality of the vector representing the token (we use $300d$ GloVe embeddings \cite{glove}).

A graph is defined as $G = (V, E)$, where $V \coloneqq$ is the set of nodes and $E \coloneqq$ the set of edges.
Each node $v_{i} \in V$, is associated with a natural language description, and we initialize them using GloVe embeddings, leading to $v_i \in \mathbb{R}^{d_v}$, with $d_v = 300$.
Similarly, we encode the edge information, with the identity of $e_{i,j}$ defined by its natural language token, represented as $e_{i,j} \in \mathbb{R}^{d_e}$, with $d_e = 300$ as well.

The primary aim of our interpretable approach is to intrinsically generate a subgraph $S_g$ as an explanation, effectively identifying the most salient nodes within $G$ as $V(S_g) \subset V(G)$.
A prediction is defined as $\hat{y} = f(x, x_e, q)$, and our model implicitly identifies the subgraph $S_g$ from the node features $x$, edge features $x_e$, and question features $q$.
We implement this through a trainable hard-attention node mask.

\section{Proposed Model Architecture}
\label{sec:method}
\begin{figure*}
    \centering
    \resizebox{0.99\textwidth}{!}{
        \includegraphics[width=\textwidth, trim=0cm 5cm 0cm 1cm,clip]{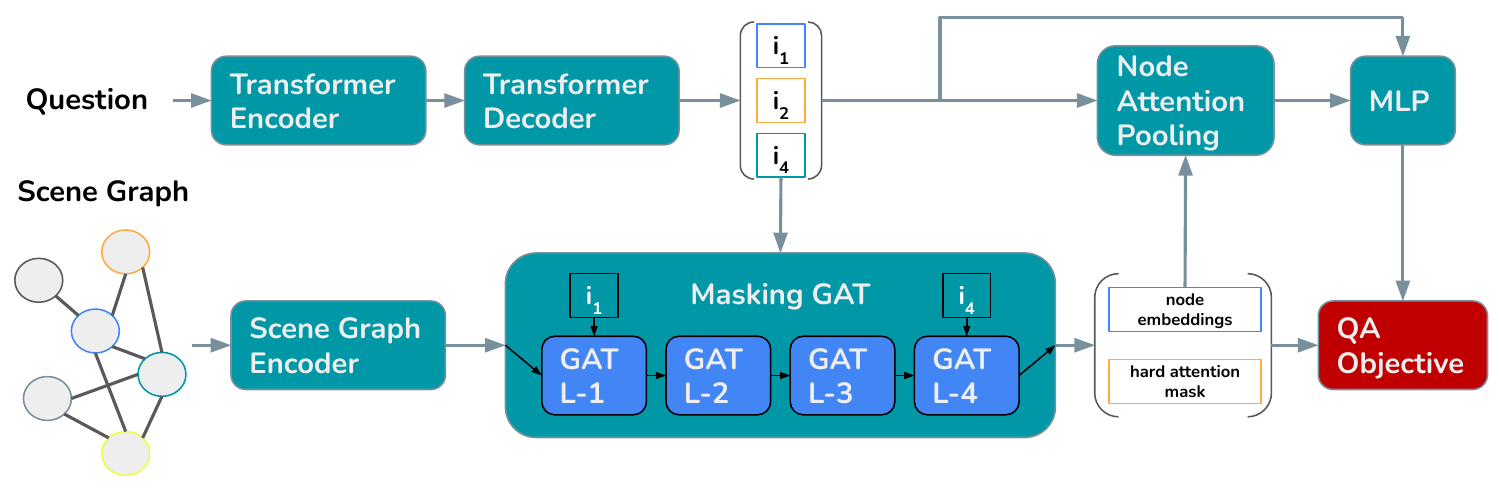}
    }
    \caption{High-level architecture of our model. The hard attention mask gets computed in \gls{gat}-layer four. Only the $top-k$ node embeddings within the hard attention mask are passed in the node attention pooling module.}
    \label{fig:arch}
\end{figure*}

We present the fundamental components of our approach in \cref{fig:arch} and \cref{alg:pipeline_model}, providing an overview of the sub-modules within our system and the algorithmic procedures involved.
At the center of our model, we employ a \gls{gat}, aimed to construct an increasingly inherently interpretable system.
This \gls{gat} learns to intrinsically generate a relevant subgraph from the input graph that contains nodes, which are most relevant for the answer finding process.
The prediction process exclusively utilizes nodes within the extracted subgraphs.

In \cref{alg:pipeline_model}, the variable $X \in \mathbb{R}^{n \times d}$ denotes the node embedding matrix.
Correspondingly, $X_e \in \mathbb{R}^{m \times d}$ represents the edge embedding matrix.
$A$ denotes the adjacency matrix, containing the interconnections between node pairs.
Furthermore, $Q$ defines the sequence of tokens in the questions.
\begin{algorithm}
    \small
    \setstretch{1.0}
    \caption{Our model pipeline}
    \label{alg:pipeline_model}
    \begin{algorithmic}[1]
        \Require Input Data: $X, X_e, A, Q$
        \State Encode Question:
        \State \quad $Q_{\text{enc}} \gets \text{Transformer Encoder}(Q)$
        \State Decode Question into Instruction Vectors:
        \State \quad $I \gets \text{Transformer Decoder}(Q_{\text{enc})}$
        \State Global Question Representation:
        \State \quad $Q_{\text{global}} \gets \text{MLP}(I)$
        \State Encode Scene Graph:
        \State \quad $X', X_e' \gets \text{Scene Graph Encoder}(X, X_e, A)$
        \State MGAT, outputs Mask for Subgraph:
        \State \quad $X', \text{mask} \gets \text{MGAT}(X', X_e', A, I)$
        \State Apply Hard-Attention Mask:
        \State \quad $X' \gets X' \times \text{mask}$
        \State Global Graph Representation Vector:
        \State \quad $X_{\text{g}} \gets \text{Global Attention}(X', Q_{\text{global}})$
        \State Final Answer Prediction:
        \State \quad $\text{logits} \gets \text{MLP}(X_{\text{g}}, Q_{\text{global}}, X_{\text{g}} \times Q_{\text{global}})$
        \State \Return Model Output: $\text{logits}, \text{mask}$
    \end{algorithmic}
\end{algorithm}

\paragraph{Question Processing}
We process the questions $Q$ with an encoder-decoder architecture.
Both the encoder and decoder are transformer-based models \cite{transformer}, randomly initialized and not fine-tuned versions of pretrained transformer models.
The token sequence $Q$ get transferred into their respective vector representations through an Embedding layer, constructed from the dataset vocabulary, within the transformer encoder.
We utilize \citet{glove} vectors as initial vector representations of the Embedding layer.
The transformer decoder takes the encoded sequence $Q_{enc}$ as input and generates a fixed-length sequence, denoted  as instruction vectors, similar to the approach by \citet{graph_vqa}.
These instruction vectors, represented as $I$, distribute information from the textual modality (questions) to the visual modality (scene graphs representing image scenes).
The instruction vectors $I$ are flattened and projected in the hidden dimensionality using a \gls{mlp} to produce the global question vector $Q_{global}$.
This global question vector plays an important role in the final answer token prediction and guides the global attention aggregation following the \gls{gnn} processing step.
The number of instructions corresponds directly to the number of layers we use in our \gls{gnn}.

\paragraph{Scene Graph Encoding}
The scene graph encoding module functions as an interface, connecting the graph to the \gls{gnn} processing it.
This encoder translates node identities (e.g., \textit{building}) and up to three corresponding attributes (e.g., \textit{large}, \textit{grey}, \textit{square}) into vector representatins using 300-dimensional GloVe vectors, akin to the initialization of question tokens.
The node representation is the summation of these vectors.
In \cref{alg:pipeline_model}, the node's identity and attributes are contained in $X$, while edge (relation) information in encoded in $X_e$.
The bounding-box coordinates are encoded using a \gls{mlp}. Subsequently, the bounding-box vector and node representation are concatenated and projected back into the defined hidden dimensionality.

To obtain the final representations, we process both the node embeddings $X$ and edge embeddings $X_e$ through an \gls{mlp} to obtain $X^{\prime}$ and $X^{\prime}_e$.
In the case of edge embeddings, we additionally concatenate the connected nodes prior to the \gls{mlp}.

\paragraph{Masking Graph Attention Network}
Our proposed approach aims to constrain the model's usage of a subset of nodes of the original input graph, referred to as \emph{subgraph}, for making predictions.
This subgraph should be the only information that the model has access to when generating answers for given questions.
Hence, we want to discretely sample from the input graph $\mathcal{G}$, which is typically not easily differentiable using gradient-based backpropagation methods.
To differentiate the aforementioned sampling step, we estimate the gradients with \gls{imle} \cite{imle}.
In our implementation, we introduce an additional hard attention mask consisting of zeros and ones for nodes, i.e. a binary mask.
We extent the \gls{gat} and refer to it as \gls{mgat}. 

The hard attention mask is computed during message-passing in a \gls{mgat} layer.
For each node $x_i$ within the input graph $\mathcal{G}$, we compute a score $s_{i}$ using the scoring function:
\begin{equation}
 s_{i} = \sigma\left(\frac{X^{\prime}I^{T}}{\sqrt{d_x}}\right).
\end{equation}

This function calculates the scaled dot-product between node embeddings $X^{\prime}$ and the instruction vector $I$, and $d_x$ representing the dimensionality of $X^{\prime}$ and $I$.
The purpose of this step is to allow the model to learn an importance score for each node.
To obtain our $\gamma_i$, we map the scores $s_{i}$ to zero or one, which is typically non-differentiable but by incorporating \gls{imle} we can backpropagate through this discrete step.
The idea is to expand the mask on a node level to get a mask on an edge level of the graph $\mathcal{G}$.
For each $\alpha_{i,j}$ in a \gls{gat} we want to get a corresponding $\gamma_{i,j}$, which is
\begin{equation}
    \gamma_{i,j} = 
    \begin{cases}
      0, & \text{otherwise} \\
      1, & \text{if}\ s_{i}*s_{j}=1 
    \end{cases}
\end{equation}
During message passing, we zero out edges and nodes that are not part of our mask.
The new update rule is defined as:
\begin{equation}
    \mathbf{x}^{\prime}_i = \alpha_{i,i}\mathbf{\Theta}\mathbf{x}_{i} +
    \sum_{j \in \mathcal{N}(i)} \mathbf{\Theta}\mathbf{x}_{j}\alpha_{i,j}\gamma_{i,j},
    \label{eq:mgat_update_rule}
\end{equation}
Here, $\mathcal{N}(i)$ denotes all neighboring nodes $\mathbf{x}_{j}$ of node $\mathbf{x}_{i}$.
The computation of $\alpha_{i,j}$ remains unchanged.
It is worth noting that, in our experiments, the hard attention mask and its effect on message-passing are only applied in the final layer.

During backpropagation, we aggregate gradients received on edge level via summation or multiplication to estimate the gradients of our mask at a node level.

To enhance the flow of information between \gls{gnn} layers in our \gls{mgat}, we employ residual connections \cite{skipconnection} with a forget-gate mechanism, defined as $l(x) = f(x) + x$.

\paragraph{Global Attention Aggregation \& Question Answering}
The node embeddings $X^{\prime}$ of the \gls{mgat} (line none to twelve in \cref{alg:pipeline_model}) are aggregated form a unified graph embedding vector, employing a global attention mechanism driven by the question vector $Q_{global}$.
This $Q_{global}$, derived from the instruction vectors, performs a scaled dot-product with $X^{\prime}$ to obtain $\alpha_i$ scores, which are normalized using \emph{softmax}.
These $\alpha_i$ scores guide the weighted summation, resulting in a single graph embedding vector $X_g$.
To perform question answering, we take the graph embedding vector $X_g$, the $Q_{global}$ representation, and the the Hadamard product of both, concatenate them before feeding them through a \gls{mlp} to receive our final answer token logits.
We optimize the system using a cross-entropy loss and Adam-W \cite{adam-w}.

\section{Evaluation Methods}
Due to the unavailability of ground-truth explanations, we employ post-hoc \gls{xai} methods for graph neural networks as a benchmark for our intrinsically generated subgraphs.
To assess the explanations, we (i) perform a human evaluation as qualitative analysis, and (ii) introduce metrics for quantitative assessment.

\subsection{Baselines for Comparison}
We included \emph{Integrated Gradients} \cite{integrated_gradients} a gradient-based method and popular \gls{xai} method, and \emph{PGMExplainer} \cite{pgmexplainer} and \emph{GNNExplainer} \cite{gnnexplainer} as perturbation-based methods.
Additionally, we include a Random explainer as a baseline.
We leverage the implementations of \citet{graphxai}, except for the GNNExplainer\footnote{The implementation of \citet{pyg} offered more flexibility, leading to improved results.}.

GNNExplainer learns a soft-mask applied the nodes.
To ensure consistency with the number of nodes in the subgraph, we utilize the same $topK$ value that we computed for our mask.

\subsection{Human Evaluation}
In our study, we include four explainability methods: (1) our intrinsic subgraph, (2) GNNExplainer due to its notable performance in quantitative metrics (cf. \cref{sec:results:token_coo} and \cref{sec:results:rem_subg}), (3) Integrated Gradients as gradient based method, and (4) the random explainer serving as a baseline.
PGMExplainer was excluded due to GNNExplainer performing better, and both being pertubation-based methods (cf. \cref{sec:results}).

To perform pairwise comparisons among these methods, we conducted a total of six comparisons that span all four techniques.
Participants were presented with 18 randomly selected graph-question pairs.
It's worth mentioning that the image was additionally displayed solely as a reference, as the model does not utilize the image as input.
Notably, each possible method pair occurred exactly three times.
For each graph, two subgraph visualizations from different explainability methods were provided as the corresponding explanations given the question and graph.
Participants were tasked with selecting their preferred explanation, or they could opt for one of two additional choices: \emph{equally good} or \emph{equally bad}.
The latter was to be chosen when none of the explanations was deemed suitable for the given graph-question pair, while the former represented a valid choice when both explanations were deemed acceptable, with no preference between them.

To (i) minimize potential psychological biases (instances where users refrain from selecting "equally good" because they deem both explanations unsatisfactory) and (ii) collect a more comprehensive dataset, we split the \emph{equal} option into \emph{equally good} and \emph{equally bad}.
We randomize the order and orientation of the comparisons, enabling a robust evaluation of user preferences.

We choose the Bradley-Terry model \cite{bradley_terry_1, bradley_terry_2} to analyze the results, and we treated \emph{equally good} and \emph{equally bad} as ties with a score of $0.5$, given their identical nature in pairwise comparisons.

\subsection{Metrics for Subgraphs}

\paragraph{Answer and Question Token Co-occurences}
To get an estimate how well the subgraphs capture the information given by both the question and its corresponding answer, we conduct a token co-occurrences analysis.
We expect the subgraphs to include objects that are mentioned in the question, as well as the answer token (if the answer is indeed an object), to appear more frequently in the explanation subgraph.
Otherwise, the model's prediction seems implausible.

When the ground-truth answer is an object that can occur in the scene graph, we count the occurrences of the respective token within the subgraph and report the resulting percentage.
Likewise, we calculate the percentage of question tokens present in the subgraph.

\paragraph{Removing Subgraphs}
To measure the nodes' importance for predictions, we suspect the nodes included in our explanation subgraph to be crucial. 
Accordingly, we aim to \textit{remove} the subgraph and passing the remaining graph through the network to measure answer accuracy once more.

To \textit{remove} a subgraph we do not alter the structural integrity of the scene graph.
Instead, we randomize the node embeddings, i.e. for each $x_i \in X$, where $X$ is the matrix of node embeddings, we randomize $x_i$ if $\gamma_i$ equals $1$.

Furthermore, the edge embeddings are also randomized.
Specifically, each $e_{i,j}$ between nodes $x_i$ and $x_j$ also provides no information, if $\gamma_{i,j}$ equals $1$.
This process ensures that the message-passing paradigm among nodes remains undisturbed, enabling it to occur to the same extent as in the original graph.

If the system remains capable of answering questions when information from the nodes within the subgraph is absent, two potential interpretations arise: (i) either the identified subgraph may not be as relevant, or (ii) the model could be exploiting biases present in the question-graph pair.

\begin{table*}
    \centering
    \resizebox{0.99\linewidth}{!}{
    \begin{tabular}{lccccccccccc}
    \toprule
        \textbf{Method} & \textbf{All} & \textbf{Attr} & \textbf{Rel} & \textbf{Obj} & \textbf{Global} & \textbf{Cat} & \textbf{Query} & \textbf{Verify} & \textbf{Choose} & \textbf{Logical}& \textbf{Compare} \\
        \midrule
         GNNExplainer & 0.35 & \textbf{0.41} & \textbf{0.39}& 0.18& 0.5 & \textbf{0.99} & \textbf{0.48}& 0.26& 0.36& 0.26 & \textbf{0.41} \\
         
         Int. Grad. & 0.09 & 0.13& 0.13 & 0.18 & 0.0 & 0.01 & 0.10& 0.24& 0.14& 0.14& 0.0 \\
         
         Random & 0.04 & 0.12 & 0.10& 0.12 & 0.0 & 0.0 & 0.10& 0.13& 0.12 & 0.07 & 0.26 \\
         
         Ours & \textbf{0.52} & 0.35 & 0.38& \textbf{0.52} & 0.5 & 0.0 & 0.32& \textbf{0.37}& \textbf{0.38}& \textbf{0.53}& 0.33 \\
    \bottomrule
    \end{tabular}
    }
    \caption{Results of the Bradley-Terry model. Each real valued $p_i$ score for the corresponding explainability method is displayed column-wise. \emph{Int. Grad.} abbreviates Integrated Gradients.}
    \label{tab:human_eval}
\end{table*}

\section{Experimental Setup and Results}
\label{sec:results}

\subsection{Setup}
\label{sec:results:setup}
\paragraph{Resources}
We conduct experiments\footnote{Implementation details are in the \cref{app:implementation_details}} on the GQA\footnote{Additional information is in the \cref{app:gqa-details}.} dataset \cite{gqa_2} since it is designed to test a model's real-world reasoning capabilities.
It provides ground-truth scene graphs, which enable \emph{perfect sight} using graph-structured representations of images.

Due to the unavailability of the scene graphs for the test sets of GQA, we report the performance on the validation set.
This practice aligns with common conventions in the field, as observed in previous works such as \citet{graphhopper} and \citet{graph_vqa}.
Each question type in GQA corresponds to a specific \emph{structural question} type as well as a \emph{semantic question} type.
The \emph{structural question} types are:
\textbf{Verify}: yes/no questions.
\textbf{Query}: open-ended questions.
\textbf{Choose}: questions offer two alternatives to choose from.
\textbf{Logical}: involve logical inference.
\textbf{Compare}: comparison among two or more objects in the scene.
The \emph{semantic question} types include:
\textbf{Object}: existence questions.
\textbf{Attribute}: questions about properties or position of an object.
\textbf{Category}: object identification within classes.
\textbf{Relation}: questions asking about the subject or object of an relation.
\textbf{Global}: questions about general properties of a scene.

\subsection{Results}

\subsubsection{RQ1: Question Answering Performance}
In terms of question-answering performance, as indicated by answer token accuracy, presented in \cref{tab:res:qst_acc}, our approach yields competitive results. 
It slightly surpasses the \gls{gat} model of \citet{graph_vqa} ($94.79\%$ compared to $94.78\%$ accuracy).
\begin{table}[h!]
    \centering
    \begin{tabular}{lcr}
        \toprule
        \textbf{Model} & \textbf{Masks} & \textbf{QA-\%} \\
        \midrule
        GAT & - & 94.78 \\
        Graphhopper & - & 92.30 \\
        \midrule
        Ours$_{k\%=.50}$ & $(1.0, 1.0, 1.0, 0.50)$ & 93.20 \\
        Ours$_{k\%=.30}$ & $(1.0, 1.0, 1.0, 0.30)$ & 94.21 \\
        Ours$_{k\%=.25}$ & $(1.0, 1.0, 1.0, 0.25)$ & 94.72 \\
        Ours$_{k\%=.20}$ & $(1.0, 1.0, 1.0, 0.20)$ & 94.15 \\
        Ours$_{k\%=.15}$ & $(1.0, 1.0, 1.0, 0.15)$ & \textbf{94.79} \\
        Ours$_{k\%=.10}$ & $(1.0, 1.0, 1.0, 0.10)$ & 77.88 \\
        \bottomrule
    \end{tabular}
    \caption{Question answering performance. GAT by \citet{graph_vqa} and Graphhopper by \citet{graphhopper}.}
    \label{tab:res:qst_acc}
\end{table}
\cref{tab:res:qst_acc} additionally contains several \emph{topK\%} configurations.
It is important to note that the hard attention masks are exclusively computed and applied within the final layer.
The first three layers learn contextualized embeddings for all nodes. 
Only in the last layer do we introduce constraints on the message-passing to learn the hard attention mask (as indicated by the values in the \emph{Masks} column, with a value of 1.0 for the first three layers) \footnote{Please refer to \cref{app:results}, \cref{app:tab:results} for a comprehensive overview of results.}
We use the top-performing model in the remaining experiments.

\subsubsection{RQ2: Human Evaluation}
In our study, we gathered data from 16 participants aged between 20 and 59, resulting in a total of 288 data points\footnote{Additional information is available in the \cref{app:human_eval}.}.
Utilizing the Bradley-Terry model \cite{bradley_terry_1, bradley_terry_2}, we established a probabilistic model to ascertain human preferences for various explainability methods, subsequently producing a ranking.
The derived $p_i$ scores are positive real values associated with the \emph{i-th} explainability method.
These outcomes were determined both on an overall scale and specifically for each question type.
The detailed $p_i$ scores can be observed in \cref{tab:human_eval}. In this context, the scores serve as an interpretation of the relative preferences of humans.

\subsubsection{RQ3: Token Co-occurrences}
\label{sec:results:token_coo}
\paragraph{\gls{atcoo}}
The co-occurrence rates between answer tokens and graph nodes are presented in \cref{tab:ans_token_coo}.
We utilized a random subset of $10\%$ of the evaluation data to compute the results in \cref{tab:ans_token_coo} and  \cref{tab:qst_token_coo}, due to the intensive computational costs of all explainability methods.
\begin{table}[!h]
    \centering
    \begin{tabular}{lr}
        \toprule
        \textbf{Method} & \textbf{AT-COO} \\
        \midrule
        Ours$_{k\%=.15}$ & 75.15 \\
        Random$_{k\%=.15}$ & 30.59 \\
        GNNExplainer$_{k\%=.15}$ & 89.12 \\
        PGMExplainer$_{k\%=.15}$ & 22.37 \\
        Integrated Gradients$_{k\%=.15}$ &  8.14 \\
        \bottomrule
    \end{tabular}
    \caption{\gls{atcoo}, represented as a percentage of potential matches.}
    \label{tab:ans_token_coo}
\end{table}
For our method, in $75.15\%$ of cases, the intrinsic subgraph captures the node aligned with the answer token.
Notably, the GNNExplainer outperforms with a coverage of $89.12\%$, suggesting a higher precision in encompassing the answer token node. Conversely, other methods display diminished focus on the node associated with the answer token.

\paragraph{\gls{qtcoo}}
\cref{tab:qst_token_coo} presents the co-occurrence results for question tokens and their alignment with graph nodes.
The subgraph generated by our method contains $78.35\%$ of the feasible question token matches.
\begin{table}[h]
    \centering
    \begin{tabular}{lr}
        \toprule
        \textbf{Method} & \textbf{QT-COO} \\
        \midrule
        Ours$_{k\%=.15}$ & 78.35 \\
        Random$_{k\%=.15}$ & 29.79 \\
        GNNExplainer$_{k\%=.15}$ & 59.67 \\
        PGMExplainer$_{k\%=.15}$ & 24.67 \\
        Integrated Gradients$_{k\%=.15}$ & 39.95 \\
        \bottomrule
    \end{tabular}
    \caption{\gls{qtcoo}, represented as a percentage of potential matches.}
    \label{tab:qst_token_coo}
\end{table}
Other methods show a reduced frequency of including the respective question token nodes in their explanatory subgraphs.
Noteworthy among these are GNNExplainer and Integrated Gradients, with inclusion rates of $59.67\%$ and $39.95\%$, respectively.

\subsubsection{RQ3: Removing Subgraphs}
\label{sec:results:rem_subg}
\cref{tab:removing_subgraphs} presents the question-answering accuracy when removing the explanatory subgraph, based on the same $10\%$ of the evaluation data, as in \cref{sec:results:token_coo}.
\begin{table}[h]
    \centering
    \begin{tabular}{lr}
        \toprule
        \textbf{Method} & \textbf{QA-\%} \\
        \midrule
        Ours$_{k\%=.15}$ & 37.13 \\
        Random$_{k\%=.15}$ & 52.10 \\
        GNNExplainer$_{k\%=.15}$ & 33.28 \\
        PGMExplainer$_{k\%=.15}$ & 69.46 \\
        Integrated Gradients$_{k\%=.15}$ & 33.28 \\
        \bottomrule
    \end{tabular}
    \caption{Question answering performance after randomizing the important nodes of the respective methods.}
    \label{tab:removing_subgraphs}
\end{table}
A larger reduction in accuracy implies the importance of the subgraph in capturing essential graph nodes for the given question.
Notably, the highest degradations were observed for Integrated Gradients, GNNExplainer, and our method.
In contrast, PGMExplainer and Random resulted in lesser performance reductions, indicating their generated subgraphs were less important than those of the other methods.
It is worth noting, that only our approach, our variant of GNNExplainer (refer to \cref{sec:results:setup}), and PGMExplainer are restricted to leveraging just $15\%$ of the input graph.

\section{Analysis}

\paragraph{RQ1: Intrinsically Interpretable GNN}
To answer the first research question, we incorporated a discrete sampling method into the message passing mechanism of \gls{gnns}.
This was to learn a discrete, adjustable mask capable of isolating the most important subgraph in the input with respect to a given question.
Our findings underscore that, even when constraining the model to operate on a subset of nodes, it remains feasible to attain competitive performance, thereby mitigating the gap between entirely black-box models and those that are interpretable.
It is noteworthy that the $topK\%$-hyperparameter, crucial in the model configuration, yielded the highest accuracy when just $15\%$ of nodes were actively used in making predictions.

\paragraph{RQ2: Human Evaluation}
In order to address the second research question, we conducted a user study to ascertain the preferred explainability methods among human participants\footnote{Qualitative examples are in the \cref{app:qual_ex}}.
In the evaluation, explanations derived from our method were notably preferred, registering a score of $p_i=0.52$. 
The GNNExplainer ranked second with a score of $p_i=0.35$, while both the Integrated Gradients and the random explainer demonstrated lower preferences.
Analyzing the results on a question-type specific basis, divergent preferences were identified, underscoring the competitive performance between our method and GNNExplainer.
The GNNExplainer was predominant in for the question types \emph{attribute}, \emph{relation}, \emph{category}, \emph{query}, and \emph{compare}.
In contrast, our method exhibited superior performance in the \emph{object}, \emph{verify}, \emph{choose}, and \emph{logical} question types.
For the \emph{global} category, both methods achieved congruent scores.

\paragraph{RQ3: Quantitative Evaluation}
We introduced two metrics to assess the quality of the explanations, specifically the subgraphs, whether they captured crucial aspects of the questions and answers.
Firstly, token co-occurrences between either the question or answer and the graph nodes enabled us to evaluate if the model appropriately concentrated on relevant input segments.
Secondly, we measured the performance drop when the explanation subgraphs were omitted from the input graph.

\begin{table}[h]
    \centering
    \begin{tabular}{lrr}
    \toprule
        \textbf{Metric} & \textbf{Pearson} & \textbf{Spearman} \\
    \midrule
         \gls{atcoo}   & 0.84 & 0.60 \\
         \gls{qtcoo}   & 0.99 & 1.00 \\
         Subg-Rem & -0.48 & -0.32 \\
    \bottomrule
    \end{tabular}
    \caption{Pearson and Spearman correlation scores between our quantitative metrics and the outcomes of the human evaluation.}
    \label{tab:correlation}
\end{table}
To determine the effectiveness of our metrics, we computed the Pearson and Spearman correlation between the outcomes from our human assessment and our quantitative evaluations, as illustrated in \cref{tab:correlation}.
The \gls{atcoo} exhibits a strong positive Pearson correlation and a moderate Spearman correlation. 
Both Pearson and Spearman metrics for \gls{qtcoo} present a very high correlation.
Interestingly, the subgraph removal metric exhibits a mild negative correlation for both Pearson and Spearman, a trend that aligns with our expectations as a lower accuracy for this metric indicates better performance.

From these findings, it can be inferred that explainability methods with higher token co-occurrence values or lower subgraph removal performances are also generally favored by human evaluators, as supported by our human evaluation results.

\section{Conclusion}
We proposed an interpretable approach for graph-based \gls{vqa} that intrinsically generates a subgraph as an explanation during the answer prediction.
Despite utilizing only a subset of the nodes from the input graph, the model demonstrates competitive performance.
Through a human evaluation, it was discerned that our intrinsically produced explanations were more frequently favored over other state-of-the-art post-hoc explainability methods.
Moreover, our evaluation extended beyond the conventional answer token accuracy metric, leveraging token co-occurrences between the question, answer, and graph nodes and assessing the performance decreases upon removal of the subgraph.
We presented the results of these metrics for the selected post-hoc explainability methods in comparison to ours.
The quantitative measures demonstrated correlations with human evaluators, thus serving as effective metrics for the explanation's quality.
Additionally, our method's inherent capability to generate the subgraph explanation concurrently with the answer eliminates the need for an additional method with further hyperparameter tuning, reducing the computational overhead.

\section{Ethics Statement}
All subjects gave their informed consent for inclusion before they participated in the study.
We provided a detailed description of the task and research objectives and did not collect personally identifying data from any users.
All logs and survey responses are encrypted using an anonymous hash generated based on the freely chosen username, rather than the plain username.
We verified the estimated time in our pilot study to ensure the time we selected was below the median time.
All participants took part voluntarily and could stop participating at any time.

\section{Limitations}
Generally, the performance of machine learning models is dependent on the quality and quantity of data they are trained on.
Consequently, every model learns biases from the data distributions they have been exposed to during training.
This limits the applicability to real-world scenarios, which should be tested before deployment.
In our case, the model might have picked up certain biases regarding the distributions of objects, scenes, or relations among objects.
As a result, certain categories of objects and scenes might be over- or underrepresented.
While we chose scene graphs to represent images as graphs to increase the interpretability of deep learning architectures, this simplification of scenes displayed in images might lose information about tiny nuances contained in the raw image.

\section{Acknowledgement}
Funded by Deutsche Forschungsgemeinschaft (DFG, German Research Foundation) under Germany's Excellence Strategy - EXC 2075 – 390740016.
We acknowledge the support by the Stuttgart Center for Simulation Science (SimTech).

\section{Bibliographical References}
\label{reference}

\bibliographystyle{lrec-coling2024-natbib}
\bibliography{lrec-coling2024-example}

\onecolumn
\newpage
\twocolumn

\appendix
\section{Appendix}
\label{sec:appendix}

\subsection{GQA Dataset}
\label{app:gqa-details}
GQA \cite{gqa_2} is a dataset for real-world visual reasoning and compositional question answering.
The questions are created automatically based on Visual Genome \cite{visual_genome} scene graphs.
The dataset consists of $113,018$ images and $22,669,678$ questions that are grouped into different splits.
The splits are divided as follows: 70\% train, 10\% validation, 10\% test and 10\% as challenge test set.
Each image of the training and validation split has a corresponding ground-truth scene graph \cite{gqa_2}.

\subsection{Implementation Details}
\label{app:implementation_details}
We use the AdamW \cite{adam-w} optimizer with a learning rate of $1e-4$ and a weight decay factor of $1e-5$.
Additionally, we apply an exponential learning rate scheduler with an initial learning rate of $1e-6$ and a warmup duration of $15$ epochs.
Afterward, we decrease the learning by a factor of $0.98$ for each epoch.
The total number of epochs depends on the loss on the validation set, which we observed to occur after training for $60$ to $70$ epochs (i.e. we apply early stopping).
To further enhance the training procedure, we use a gradient scaler.
We train on four GPUs with a batch size of $128$ samples.
Our model has $44,945,761$ trainable parameters.
One training epoch consists of $943,000$ data instances, the validation split of $132,062$ instances.

\subsection{Human Evaluation}
\label{app:human_eval}
Our user study was performed online via a web interface.
First of all, users need to agree on a data collection policy before a few questions regarding demographic information are asked.
Afterward, the actual user study was performed.

\paragraph{Data Collection Policy}

In the following, we list the information users need to agree to participate in the user study:
\begin{itemize}
    \item \textbf{Purpose of research:} To examine preferences of different explainability methods.
    \item \textbf{What users will perform:} Users will be provided with 18 images, a corresponding question, the prediction from the model, and explanations of two explainability methods in the form of graphs.
    \item \textbf{Time required:} Participation will take approximately 5-10 minutes.
    \item \textbf{Risks:} There are no anticipated risks associated with participating in this study. The effects of participating should be comparable to those you would experience from viewing a computer monitor for 5-10 minutes and using a mouse and keyboard. 
    \item \textbf{Limitations:} This task is suitable for all people who can read from and input text into a computer. 
    \item \textbf{Confidentiality:} Your participation in this study will remain confidential. Your responses will be assigned a code number. You will be asked to provide your MechanicalTurk ID, but this will not be stored, but rather converted to an anonymous hashed ID. You will be asked to provide your age and gender and previous experience with chatbots/business travel. Throughout the experiment, we may collect data such as your textual input, and your feedback in the form of a questionnaire. The records of this study will be kept private. In any sort of report we make public we will not include any information that will make it possible to identify you. Research records will be kept in a locked file; only the researchers will have access to the records.
    \item \textbf{Participation and Withdrawl:} Your participation in this study is voluntary, and you may withdraw at any time.
    \item \textbf{Data Regulation:} Your data will be processed for the following purposes: (1) Analysis of the respondents' evaluations of the dialog and their experience, (2) analysis of potential influencing factors for individual behavior of the participants in the interaction with the dialog system, and (3) scientific publication based on the results of the above analyses.
\end{itemize}
\begin{figure}
    \centering
    \resizebox{0.99\linewidth}{!}{
    \includegraphics{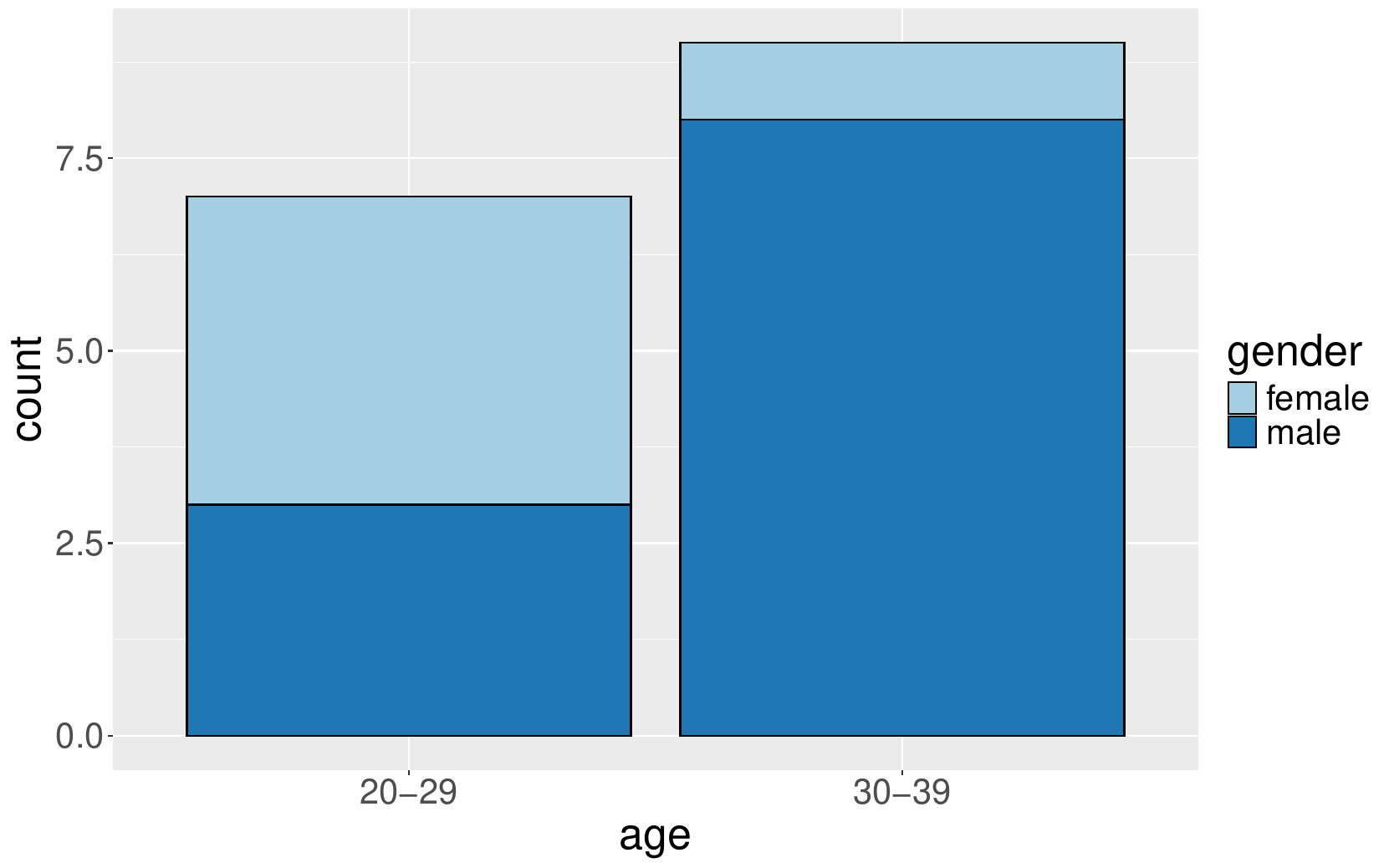}
    }
    \caption{Age and gender distribution of our participants in the user study.}
    \label{fig:age-gender}
\end{figure}
\begin{figure}[!htb]
    \centering
    \resizebox{0.99\linewidth}{!}{
    \includegraphics{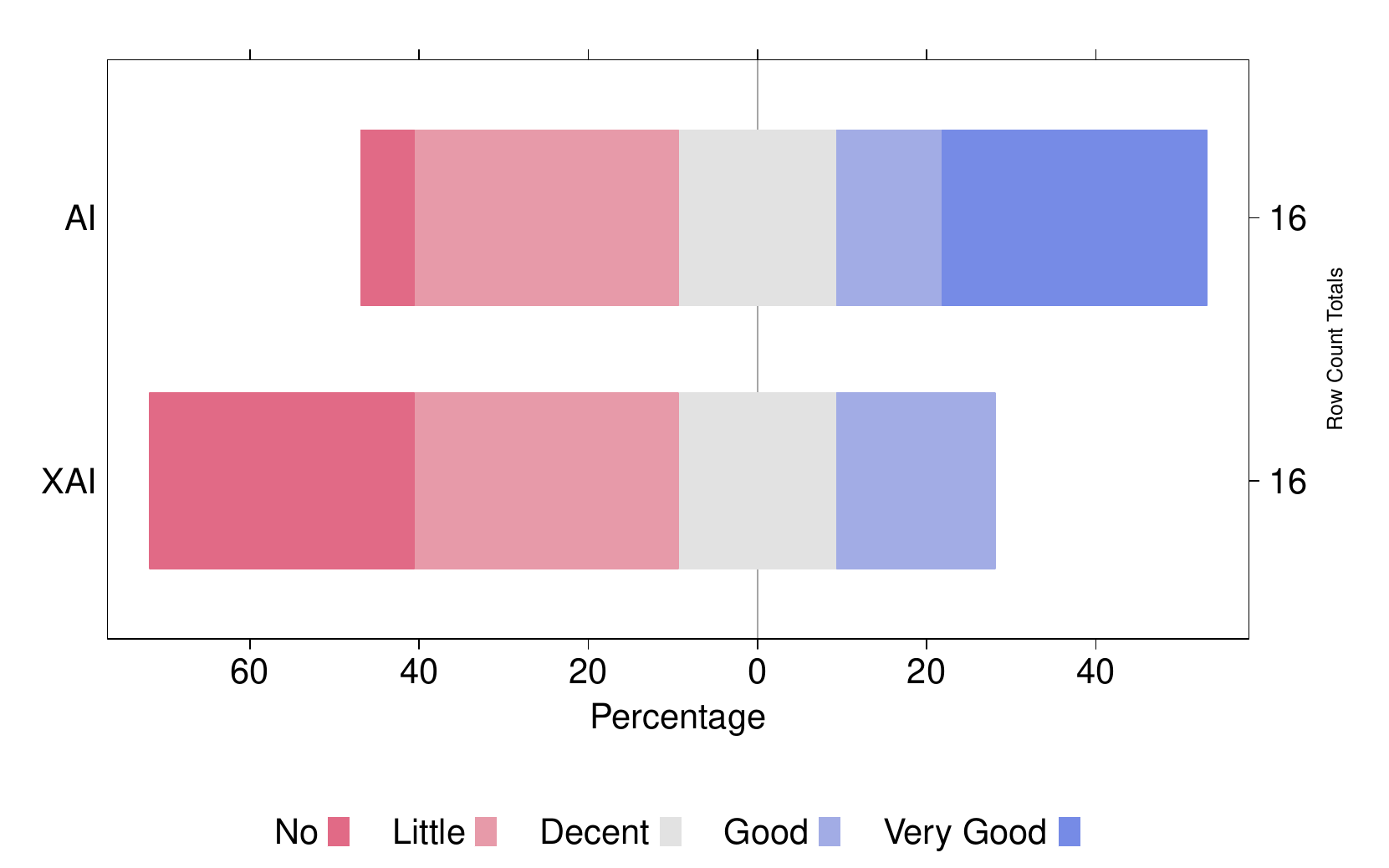}
    }
    \caption{Results of the self-assessment regarding the general topics of artificial intelligence (AI) and explainable artificial intelligence (XAI).}
    \label{fig:likert-ml-xai}
\end{figure}

\paragraph{Demographic Information}
We asked each participant about the gender they identified as, given a set of three options: female, male, and other.
Participants were given the option to select their age category from the following ranges:
\begin{itemize}
    \item Less than 20
    \item 20 to 29
    \item 30 to 39
    \item etc.
\end{itemize}
The gender and age distributions are displayed in \cref{fig:age-gender}.
The study participants were in the range from 20 to 39 with slightly more male users.

We asked two more questions about the participants' general understanding of machine learning (ML) or artificial intelligence (AI) and general understanding of the field around explainable artificial intelligence (XAI).
\cref{fig:likert-ml-xai} contains the results, measured by a Likert scale.
Our participants range from no understanding of ML to a very good understanding, contrary to the understanding in XAI where most participants have little to no knowledge about.

\paragraph{Explainability User Study}
Each participant receives 18 images accompanied by two scene graphs with the respective highlighted subgraph that serves as an explanation of the model's prediction.
The image is displayed as a reference only, it is communicated to the users that the image was not used by the model to predict an answer.
Next to the original image, we displayed the question, the predicted answer, and the ground-truth label.
We mention that some questions might be ambiguous and that this should not be evaluated or taken into account when evaluating the explanations.

We further describe that we display the corresponding graphs below the original image.
The whole graph is input to the model alongside the question. 
The graph itself (all nodes, green and blue nodes combined) might not be a perfect representation of the image. 
Nodes, which represent objects in the image, might be missing, or the annotation (the label/name) might be misleading. 
We displayed the edges between nodes in the visualization of the graph, but we excluded the annotation (the name of the relation). 
Edges represent relations between objects, e.g. \emph{a man holding a racket} would result in two nodes, \emph{man} and \emph{racket}, and one edge (relation) \emph{holding} between them.

We always perform a pair-wise comparison between two explainability methods, users can find their explanations next to each other.
All nodes colored in green are part of the subgraph that represents the explanation. 
All nodes colored in blue are excluded, so they are not part of the explanation.
To judge which explanation users prefer, they should take the question and answer into account, and evaluate if the nodes in green form a more valid explanation than the other explanation.
Some explanation methods include more graph nodes in the explanations, while others tend to have fewer nodes included.

\onecolumn
\subsection{Extended Results}
\label{app:results}
We provide the results from \cref{sec:results} for all \emph{top-k} values of \cref{tab:res:qst_acc} in \cref{app:tab:results}.
We report the question answering accuracy as \textbf{QA}, the answer token co-occurrences as \textbf{AT-COO}, the question token co-occurrences as \textbf{QT-COO}, and the question answering accuracy after removing the explanation subgraph as \textbf{QA-SubG}.
\begin{table*}[!htb]
    \centering
    \begin{tabular}{lccccccccc} 
    \toprule
         \textbf{Method} & \textbf{TopK\%} & \textbf{QA} &  \textbf{AT-COO} &  \textbf{QT-COO} &  \textbf{QA-SubG} \\
    \midrule
    \midrule
        Random & .15 & 94.79 & 30.59 & 29.79 & 52.10  \\
         PGMExplainer & .15 & 94.79 & 22.37 & 24.67 & 69.46 \\
         Integrated Gradients & .15 & 94.79 & 8.14 & 39.95 &  33.28  \\
         GNNExplainer & .15 & 94.79 & 89.12 & 59.67 & 33.28 \\
         Ours & .15 & 94.79 & 75.15 & 78.35 & 37.13 \\
    \midrule
        Random & .20 & 94.15 & 28.35 & 27.88 & 51.07 \\
        PGMExplainer & .20 & 94.15 & 26.29 & 29.94 & 70.29 \\
        Integrated Gradients & .20 & 94.15 & 8.76 & 32.78 & 30.87 \\
        GNNExplainer & .20 & 94.15 & 91.24 & 50.66 & 30.87 \\
        Ours & .20 & 94.15 & 33.51 & 80.68 & 45.63 \\
    \midrule
        Random & .25 & 94.72 & 30.41 & 33.63 & 52.10 \\
        PGMExplainer & .25 & 94.72 & 34.34 & 39.62 & 69.46 \\
        Integrated Gradients & .25 & 94.72 & 7.14 & 40.06 &  33.28 \\
        GNNExplainer & .25 & 94.72 & 92.99 & 63.60 &  33.28 \\
        Ours & .25 & 94.72 & 81.26 & 79.20 &  37.13 \\
    \midrule
        Random & .30 & 94.21 & 30.46 & 29.53 & 48.58 \\
        PGMExplainer & .30 & 94.21 & 37.93 & 38.23 & 61.74 \\
        Integrated Gradients & .30 & 94.21 & 6.90 & 36.00 & 34.82 \\
        GNNExplainer & .30 & 94.21 & 94.83 & 61.42 & 34.82 \\
        Ours & .30 & 94.21 & 78.74 & 89.11 & 39.88 \\
    \midrule
        Random & .50 & 93.20 & 31.28 & 29.82 & 48.49 \\
        PGMExplainer & .50 & 93.20 & 39.89 & 38.92 & 58.48 \\
        Integrated Gradients & .50 & 93.20 & 7.33 & 32.62 & 33.15 \\
        GNNExplainer & .50 & 93.20 & 94.68 & 61.70 & 33.15 \\
        Ours & .50 & 93.20 & 74.81 & 88.50 & 35.37 \\
    \bottomrule
    \end{tabular}
    \caption{\textbf{AT-COO} abbreviates answer-token co-occurrence.
    \textbf{QT-COO} abbreviates question-token co-occurrence.
    \textbf{QA-SubG} refers to the question answering accuracy, when the explanation subgraph is removed.}
    \label{app:tab:results}
\end{table*}
We find that the results for the methods \emph{Random}, \emph{PGMExplainer}, and \emph{Integrated Gradients} stay roughly the same across different \emph{tok-k} sized models.
However, increasing the subgraph size with the number of nodes included (determined by the \emph{top-k} factor), the \gls{atcoo} and \gls{qtcoo} increase, while the QA-SubG remains roughly the same.
Since we match the number of nodes included from the soft mask learned by \emph{GNNExplainer} with the number of nodes received by our \emph{top-k} factor, this effect is unsurprising.
Nevertheless, it is worth noting that the QA-SubG accuracy drops by a similar order of magnitude when we learn to identify a subgraph with a size of \emph{tok-k}$=.15$ or \emph{tok-k}$=.5$, highlighting the importance of nodes included in the subgraphs.

\newpage
\subsection{Qualitative Examples}
\label{app:qual_ex}
\def\figsize{0.495}
\begin{figure*}[htb]
    \centering
    \begin{subfigure}{\figsize\linewidth}
        \includegraphics[width=\textwidth, trim=0.4cm 0.4cm 0.4cm 0.4cm,clip]{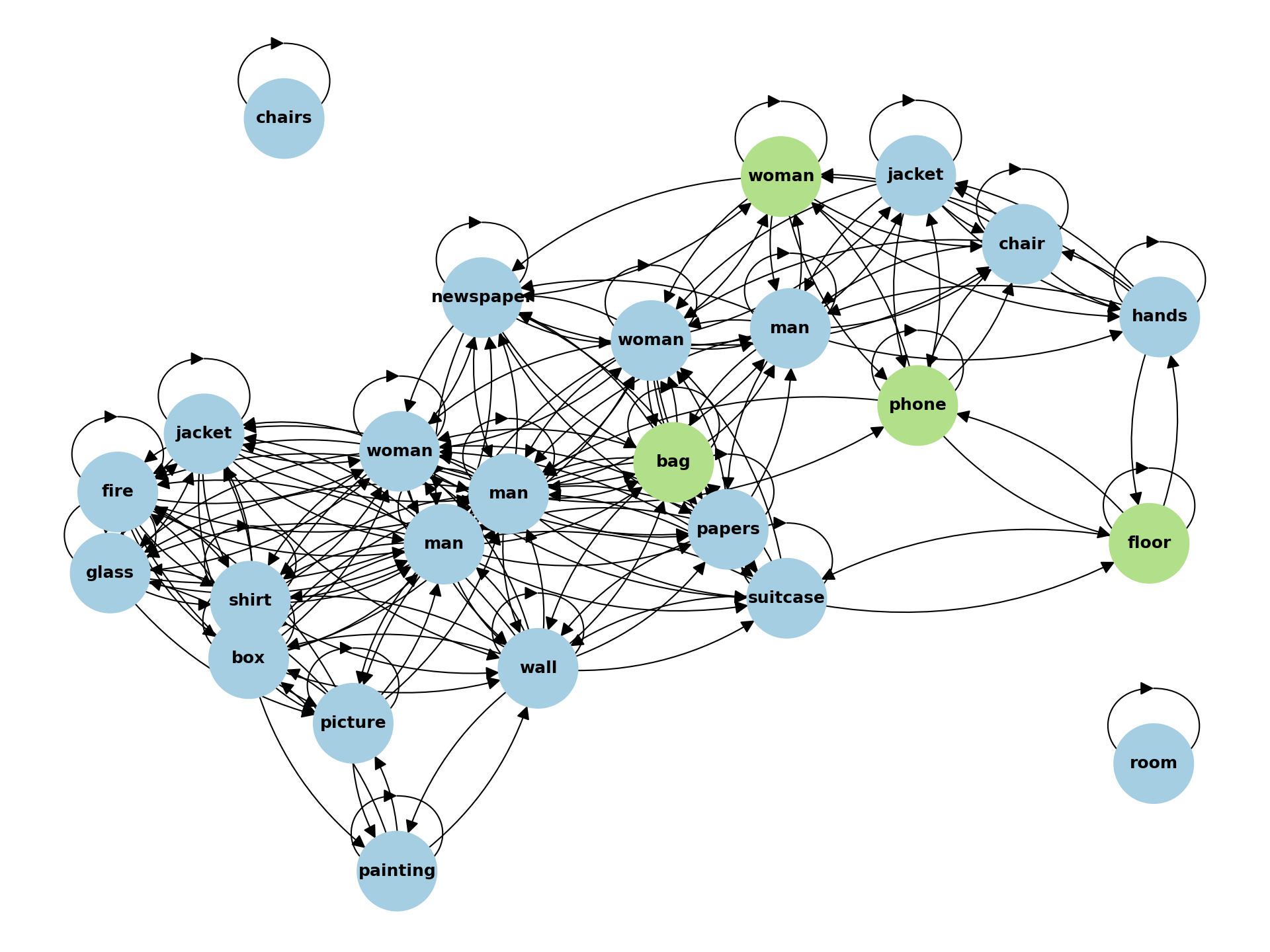}
        \caption{I-MLE.}
    \end{subfigure}
    \begin{subfigure}{\figsize\linewidth}
        \includegraphics[width=\textwidth, trim=0.4cm 0.4cm 0.4cm 0.4cm,clip]{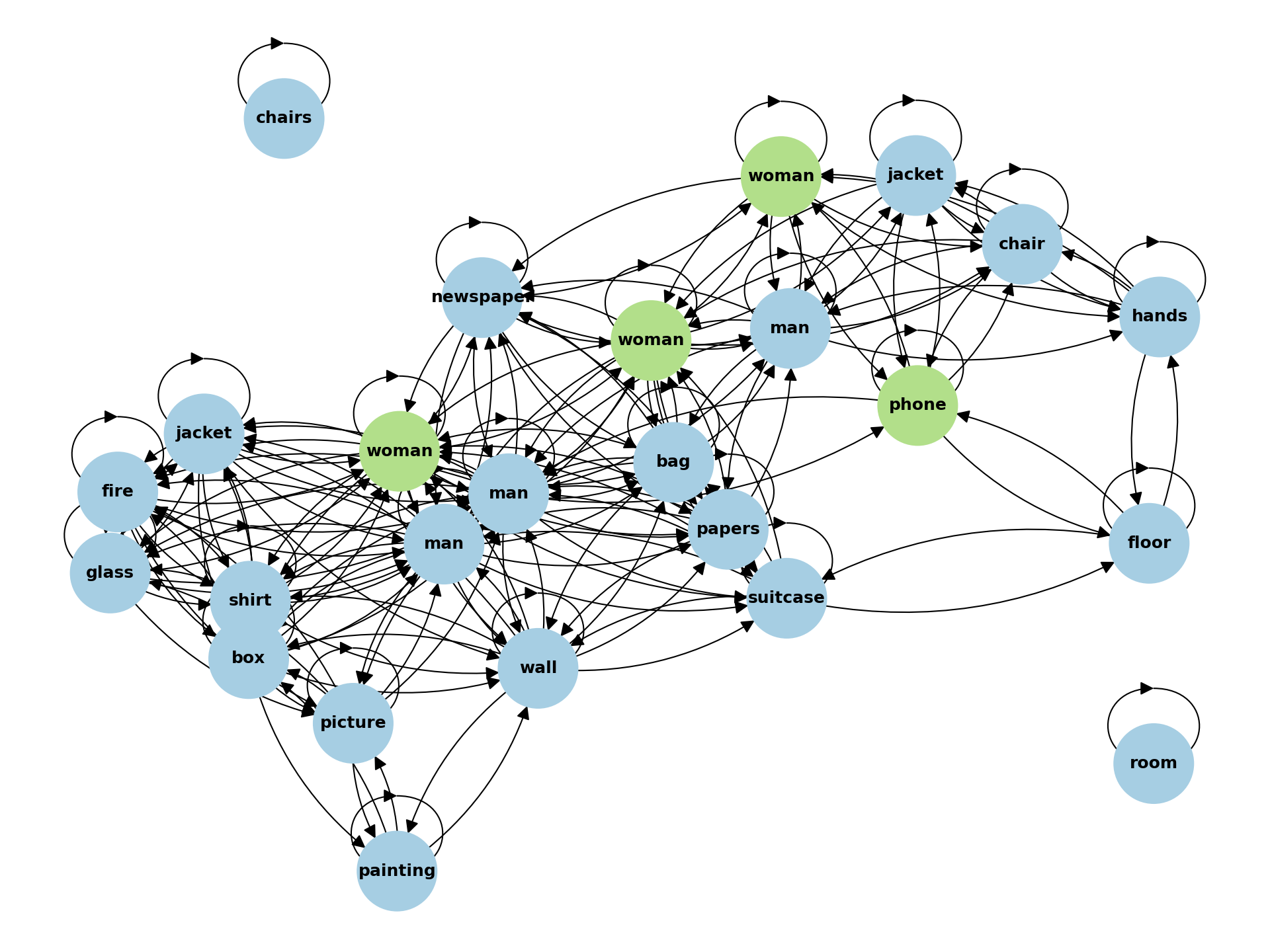}
        \caption{GNNExplainer.}
    \end{subfigure}
    \begin{subfigure}{\figsize\linewidth}
        \includegraphics[width=\textwidth, trim=0.4cm 0.4cm 0.4cm 0.4cm,clip]{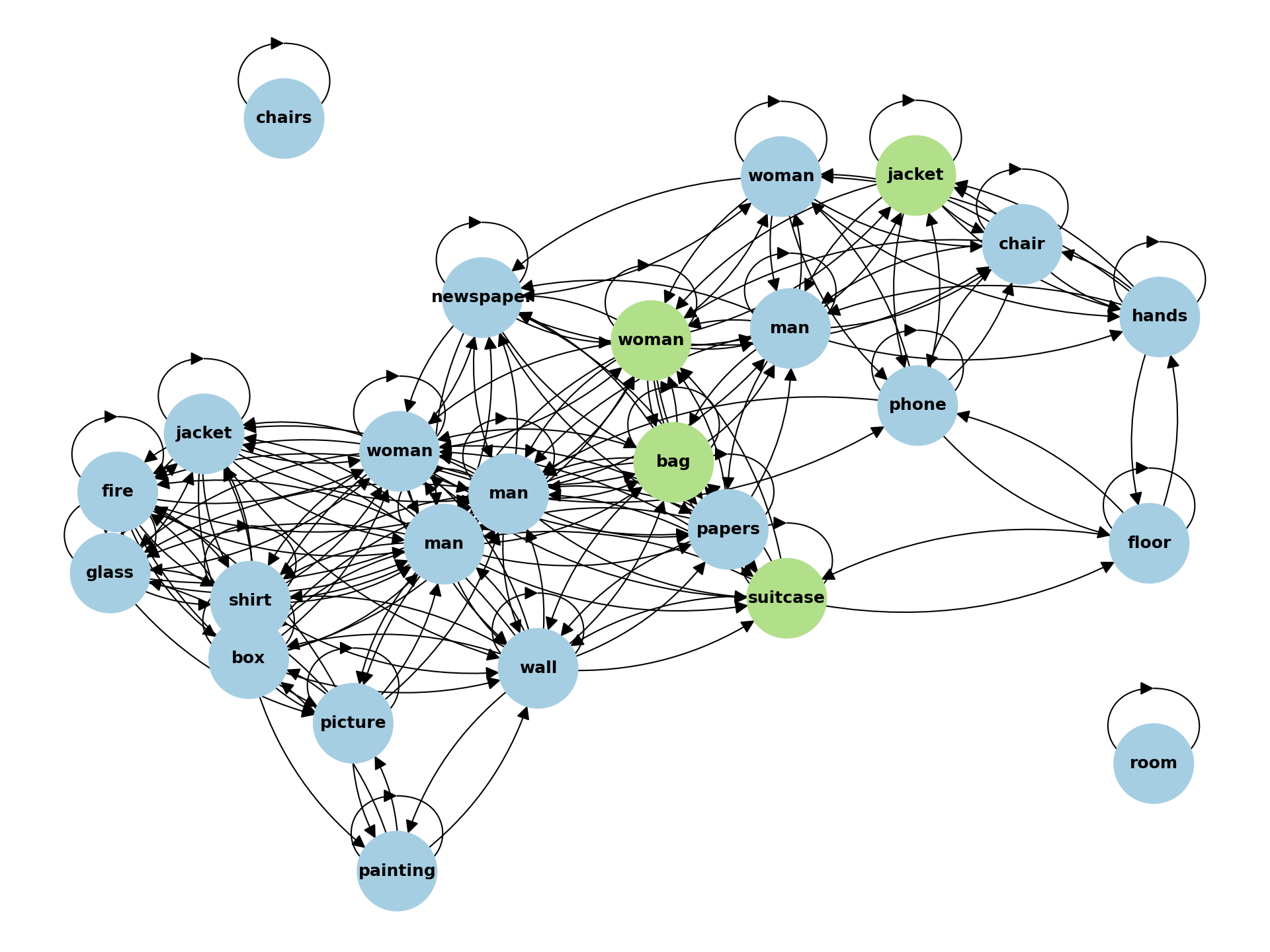}
        \caption{Random.}
    \end{subfigure}
    \begin{subfigure}{\figsize\linewidth}
        \includegraphics[width=\textwidth, trim=0.4cm 0.4cm 0.4cm 0.4cm,clip]{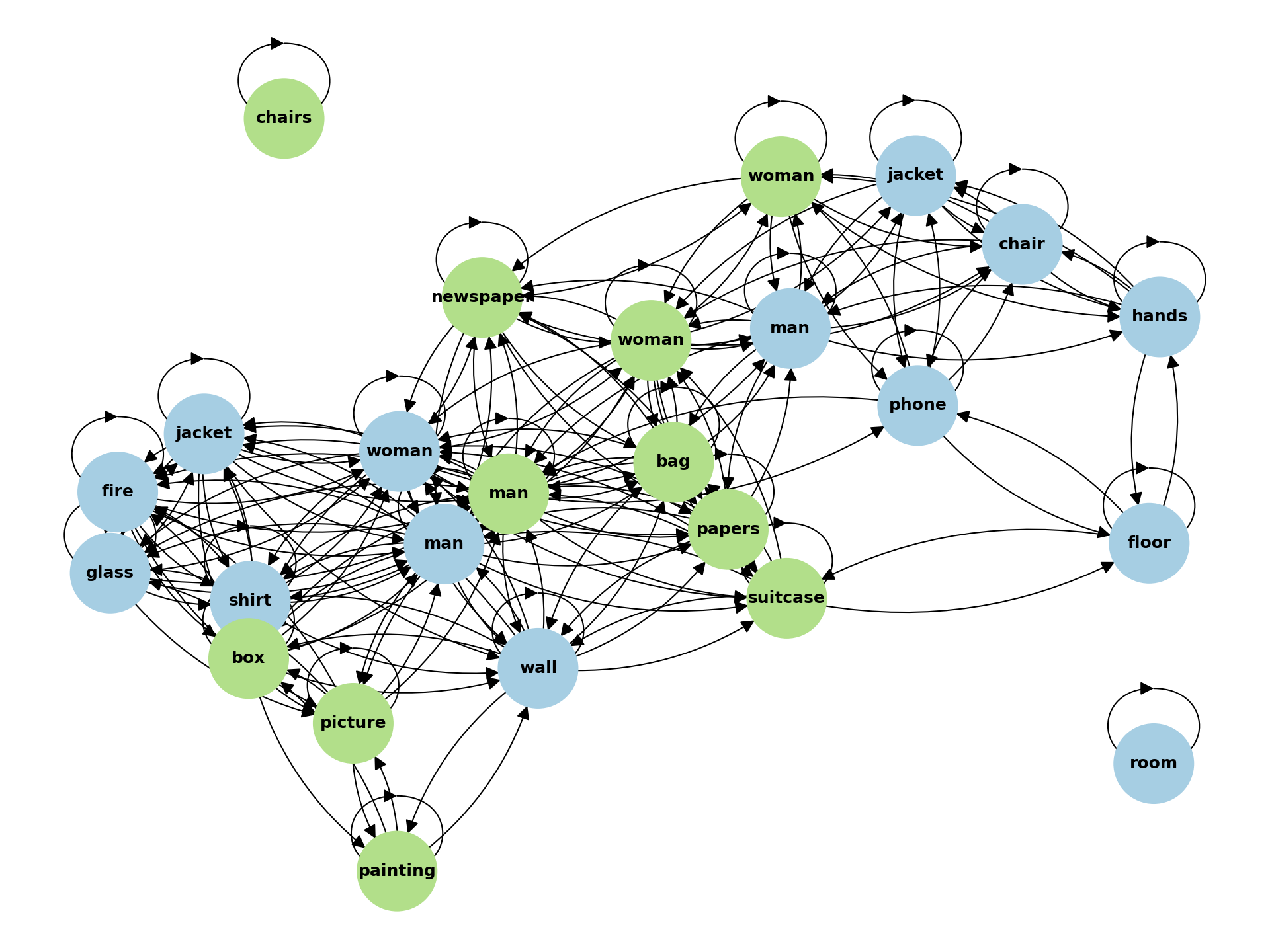}
        \caption{Integrated-Gradients.}
    \end{subfigure}
    \caption{Graph and image for question Id: 17745707. Question: \emph{Is the woman to the left or to the right of the phone?}Prediction: \emph{left}. Ground-truth answer: \emph{left}. Semantic type: \emph{relation}. Structural type: \emph{choose}.}
    \label{fig:example_1}
\end{figure*}

\begin{figure*}[htb]
    \centering
    \begin{subfigure}{\figsize\linewidth}
        \includegraphics[width=\textwidth, trim=0.4cm 0.4cm 0.4cm 0.4cm,clip]{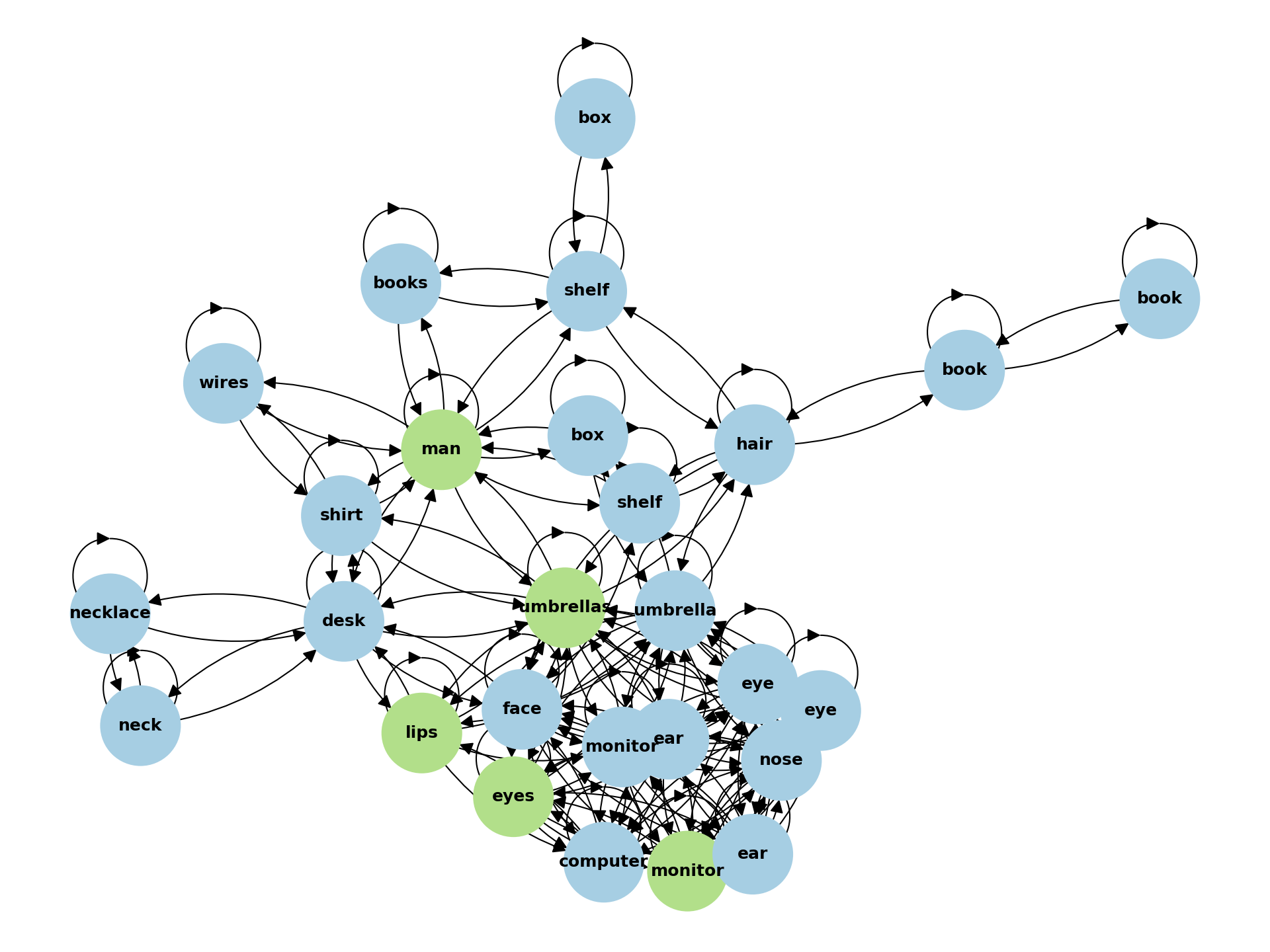}
        \caption{I-MLE.}
    \end{subfigure}
    \begin{subfigure}{\figsize\linewidth}
        \includegraphics[width=\textwidth, trim=0.4cm 0.4cm 0.4cm 0.4cm,clip]{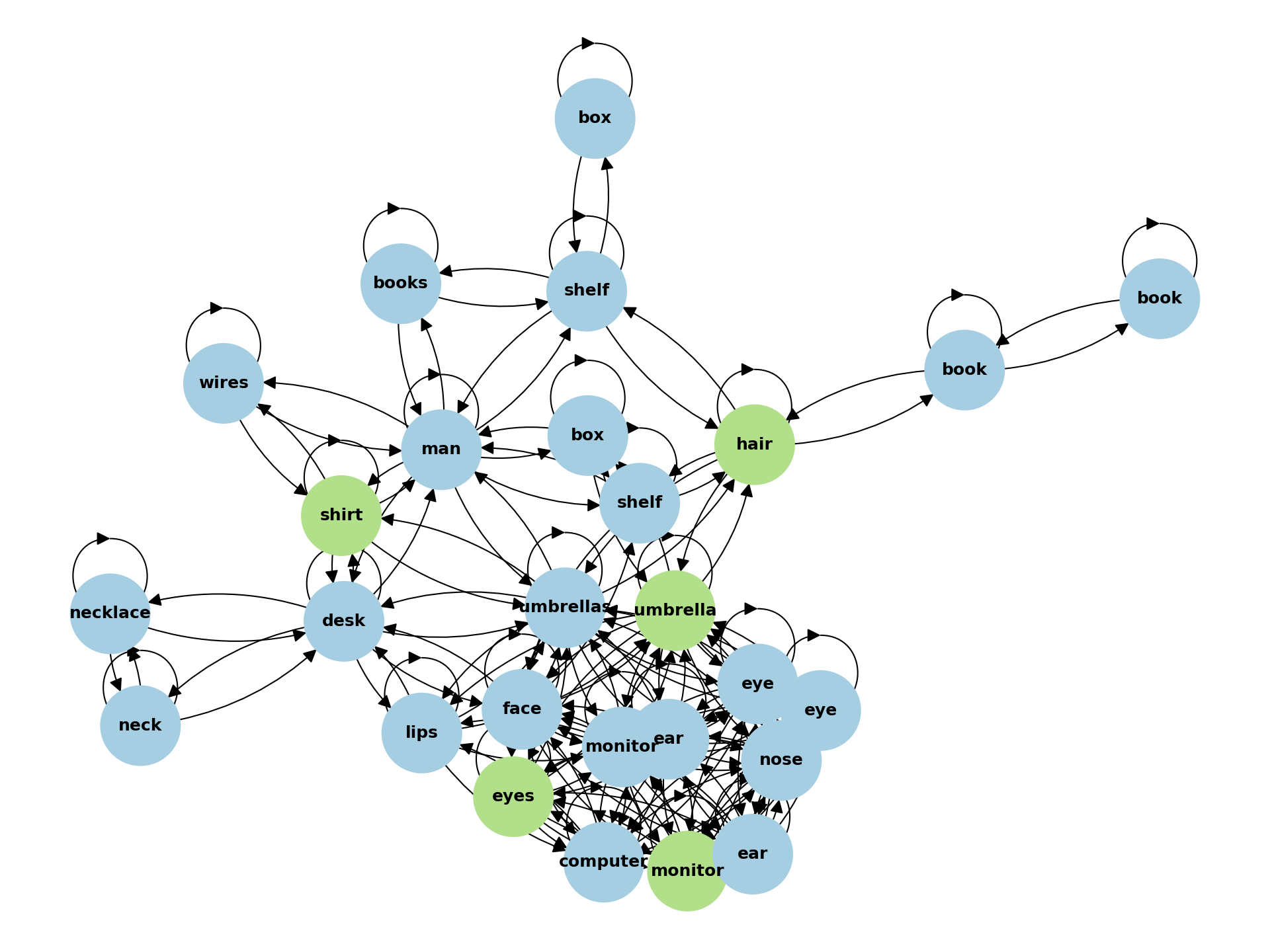}
        \caption{GNNExplainer.}
    \end{subfigure}
    \begin{subfigure}{\figsize\linewidth}
        \includegraphics[width=\textwidth, trim=0.4cm 0.4cm 0.4cm 0.4cm,clip]{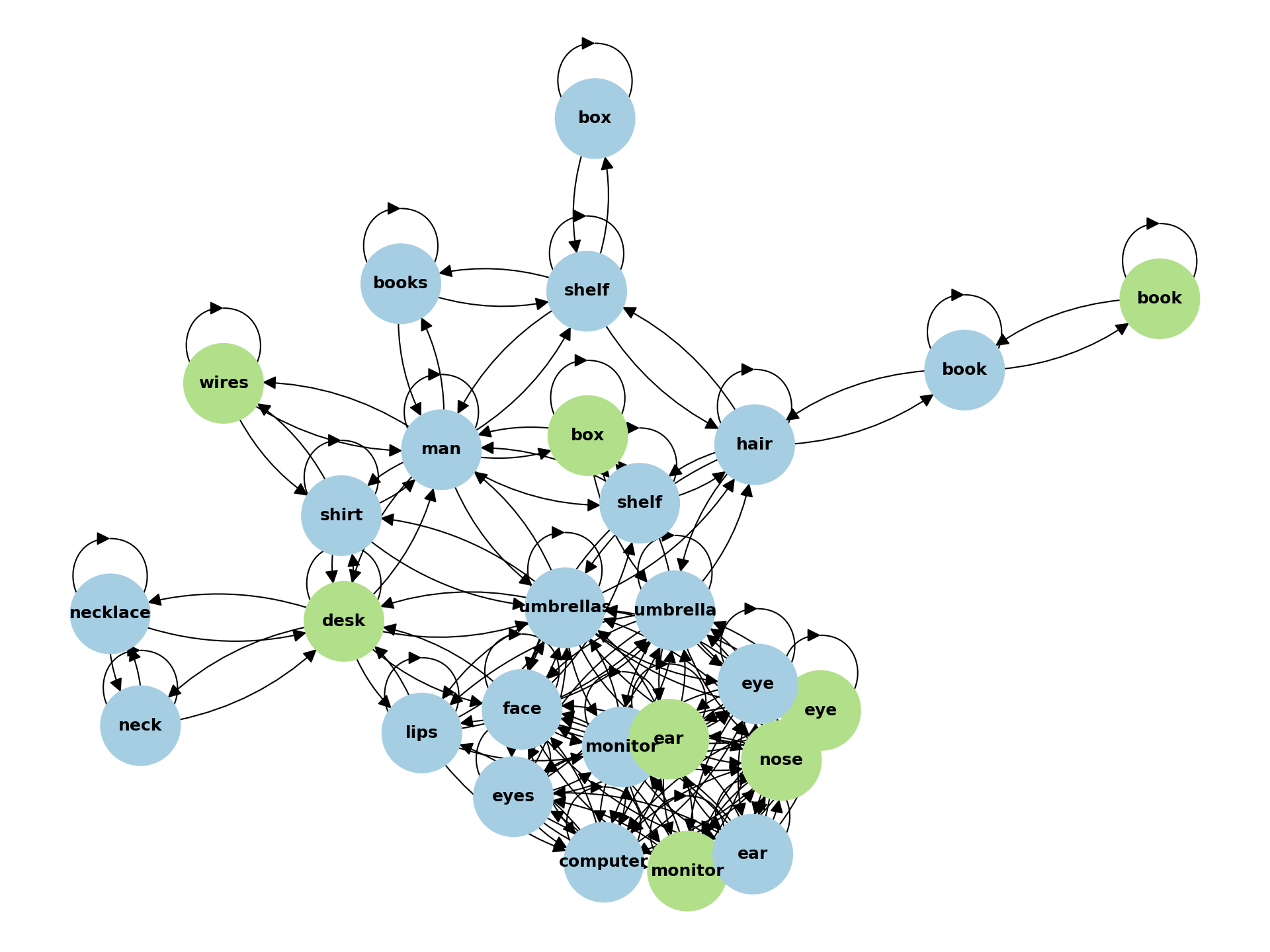}
        \caption{Random.}
    \end{subfigure}
    \begin{subfigure}{\figsize\linewidth}
        \includegraphics[width=\textwidth, trim=0.4cm 0.4cm 0.4cm 0.4cm,clip]{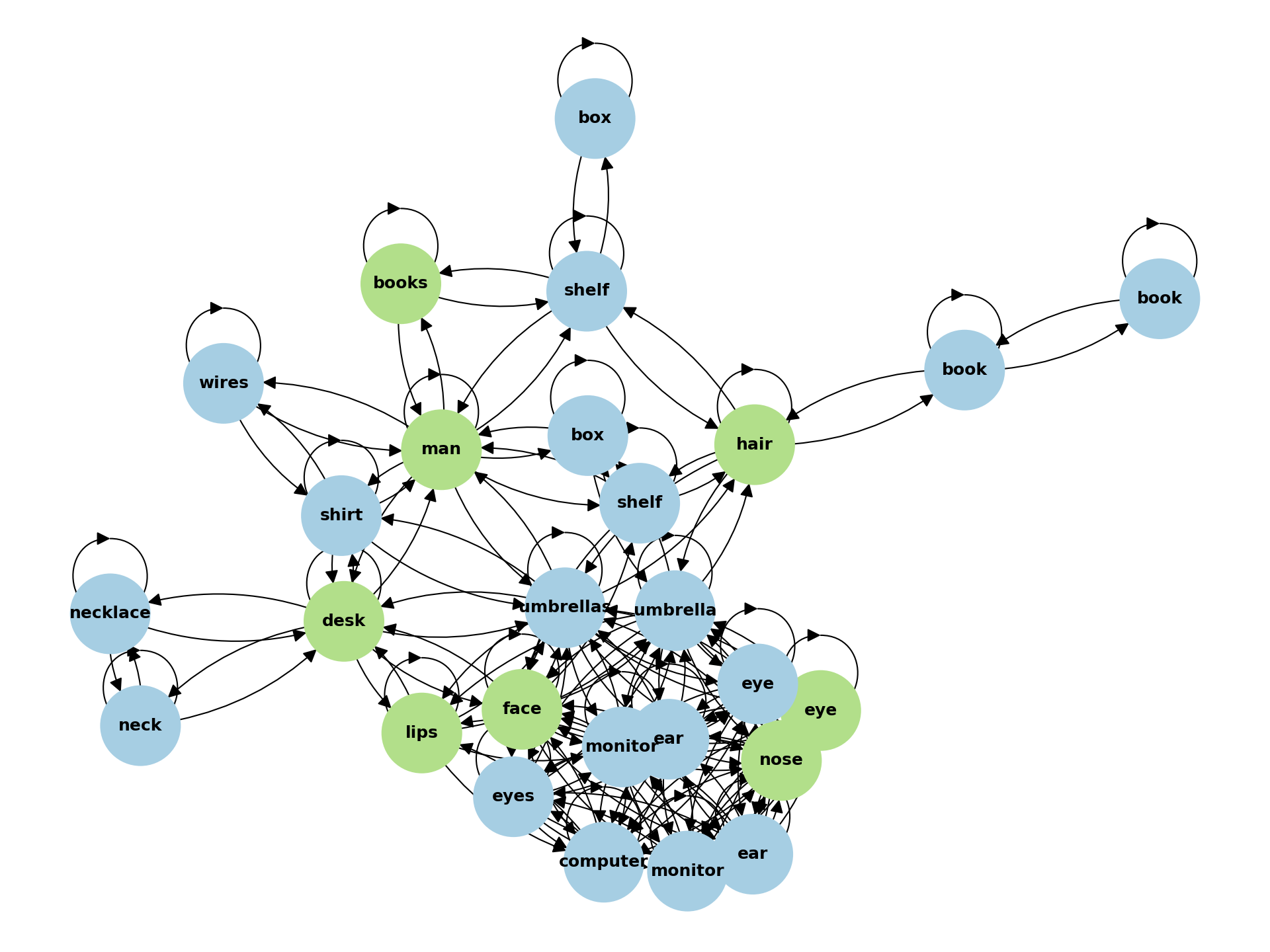}
        \caption{Integrated-Gradients.}
    \end{subfigure}
    \caption{Graph and image for question Id: 17267496. Question: \emph{Are his eyes large and green?} Prediction: \emph{no}. Ground-truth answer: \emph{no}. Semantic type: \emph{attribute}. Structural type: \emph{logical}.} 
    \label{fig:example_2}
\end{figure*}

\begin{figure*}[htb]
    \centering
    \begin{subfigure}{\figsize\linewidth}
        \includegraphics[width=\textwidth, trim=0.4cm 0.4cm 0.4cm 0.4cm,clip]{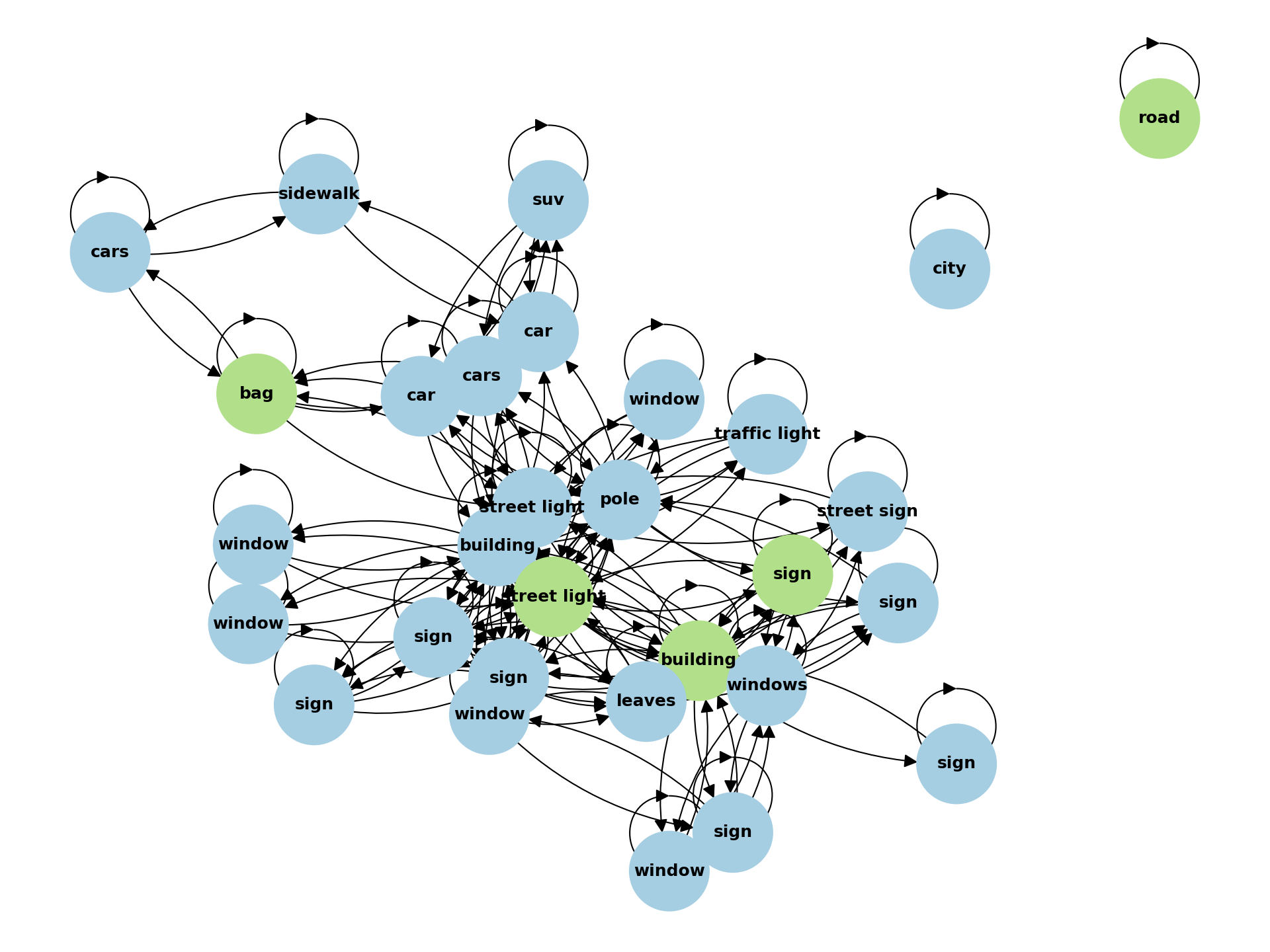}
        \caption{I-MLE.}
    \end{subfigure}
    \begin{subfigure}{\figsize\linewidth}
        \includegraphics[width=\textwidth, trim=0.4cm 0.4cm 0.4cm 0.4cm,clip]{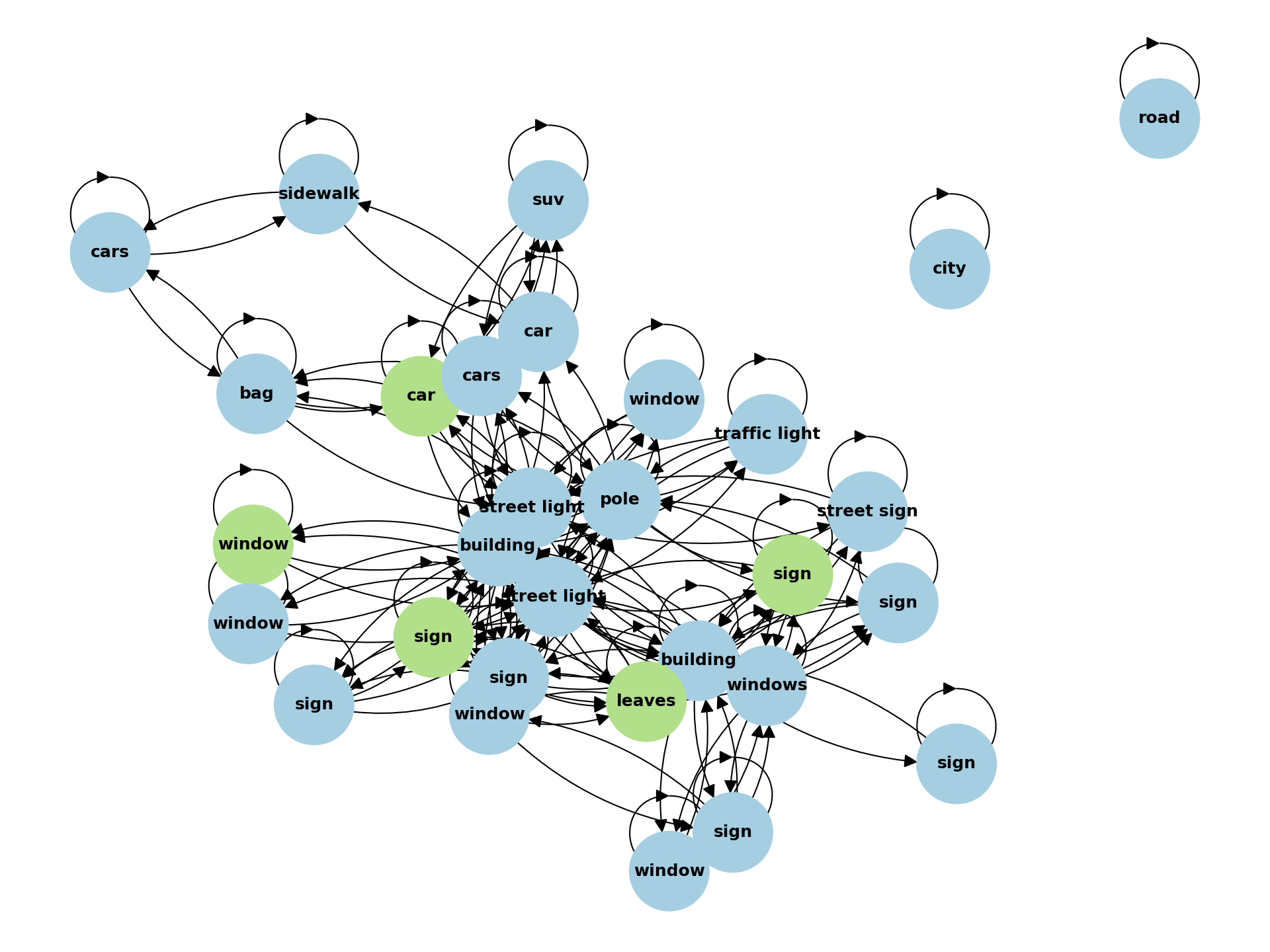}
        \caption{GNNExplainer.}
    \end{subfigure}
    \begin{subfigure}{\figsize\linewidth}
        \includegraphics[width=\textwidth, trim=0.4cm 0.4cm 0.4cm 0.4cm,clip]{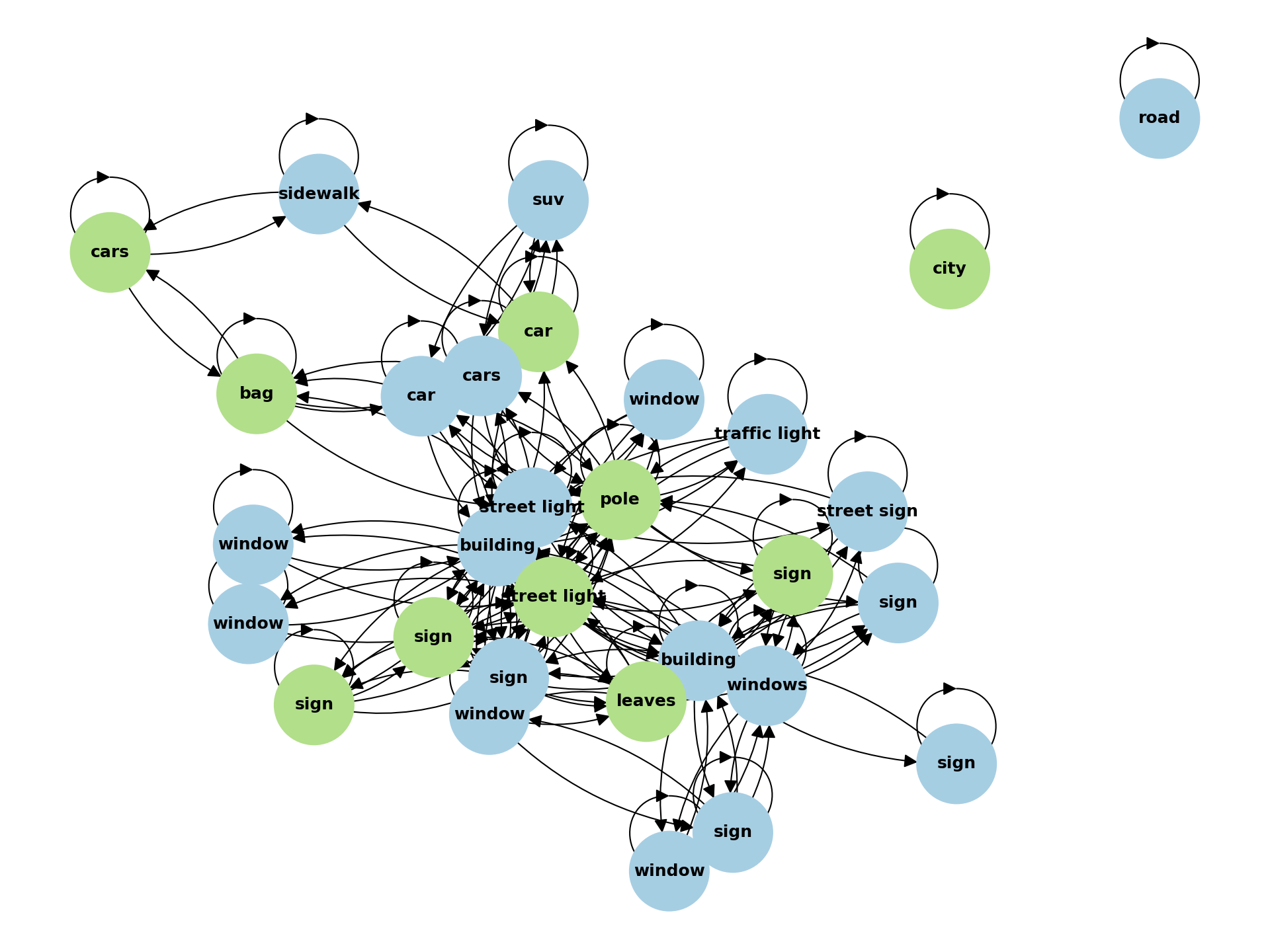}
        \caption{Random.}
    \end{subfigure}
    \begin{subfigure}{\figsize\linewidth}
        \includegraphics[width=\textwidth, trim=0.4cm 0.4cm 0.4cm 0.4cm,clip]{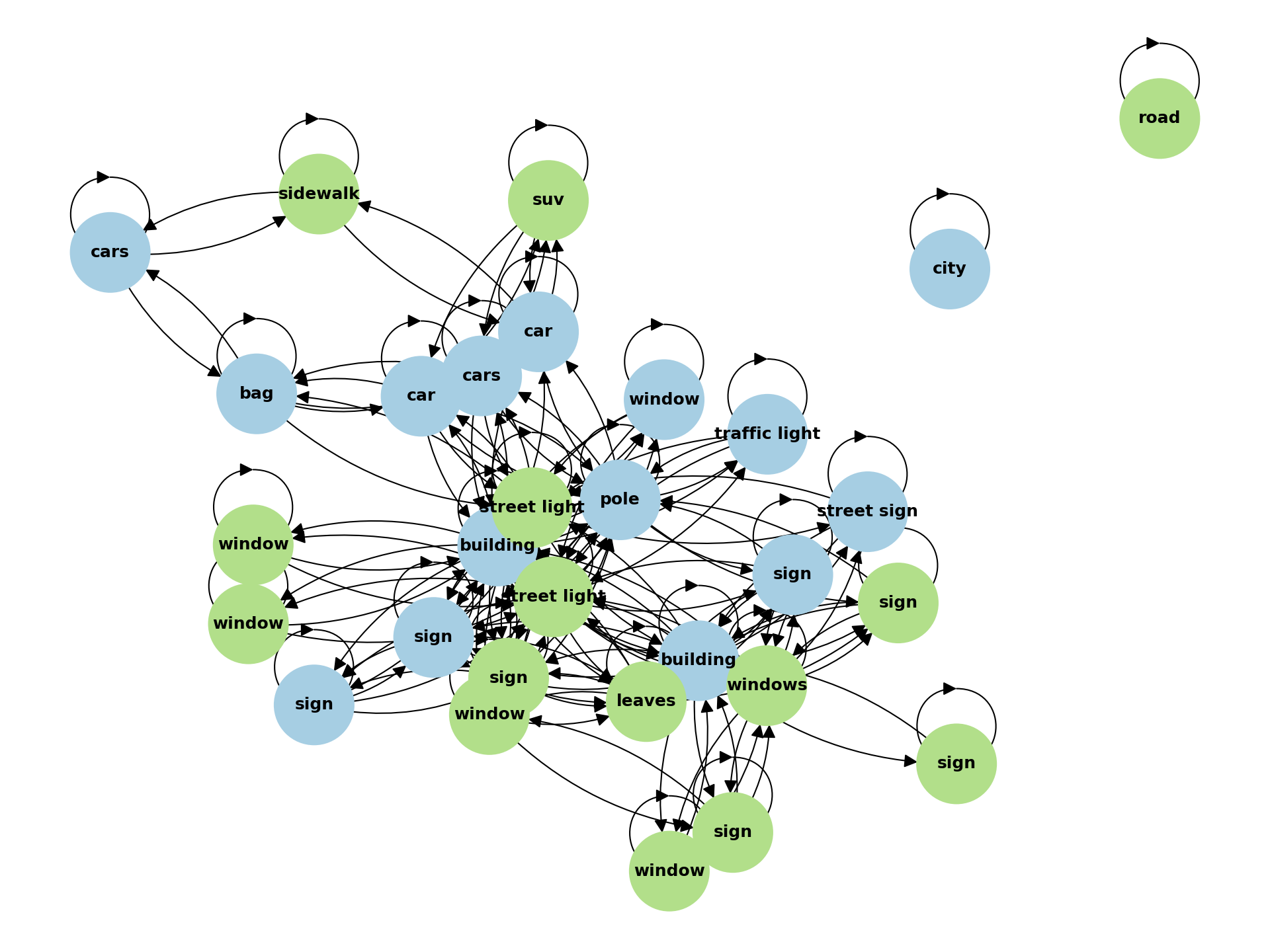}
        \caption{Integrated-Gradients.}
    \end{subfigure}
    \caption{Graph and image for question Id: 07339770. Question: \emph{Are there either pizza trays or hand soaps in the image?} Prediction: \emph{no}. Ground-truth answer: \emph{no}. Semantic type: \emph{object}. Structural type: \emph{logical}.}
    \label{fig:example_3}
\end{figure*}

\begin{figure*}[htb]
    \centering
    \begin{subfigure}{\figsize\linewidth}
        \includegraphics[width=\textwidth, trim=0.4cm 0.4cm 0.4cm 0.4cm,clip]{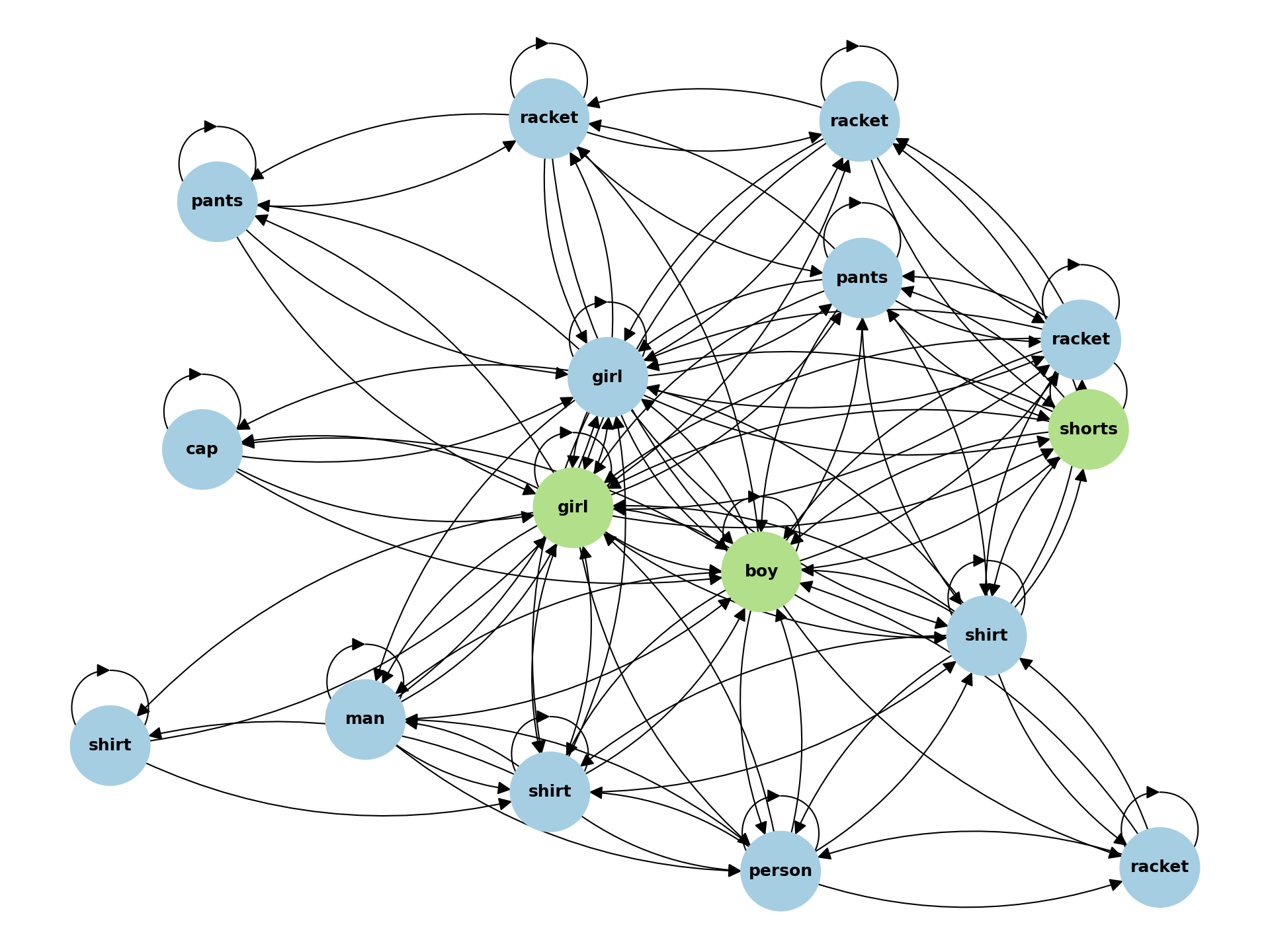}
        \caption{I-MLE.}
    \end{subfigure}
    \begin{subfigure}{\figsize\linewidth}
        \includegraphics[width=\textwidth, trim=0.4cm 0.4cm 0.4cm 0.4cm,clip]{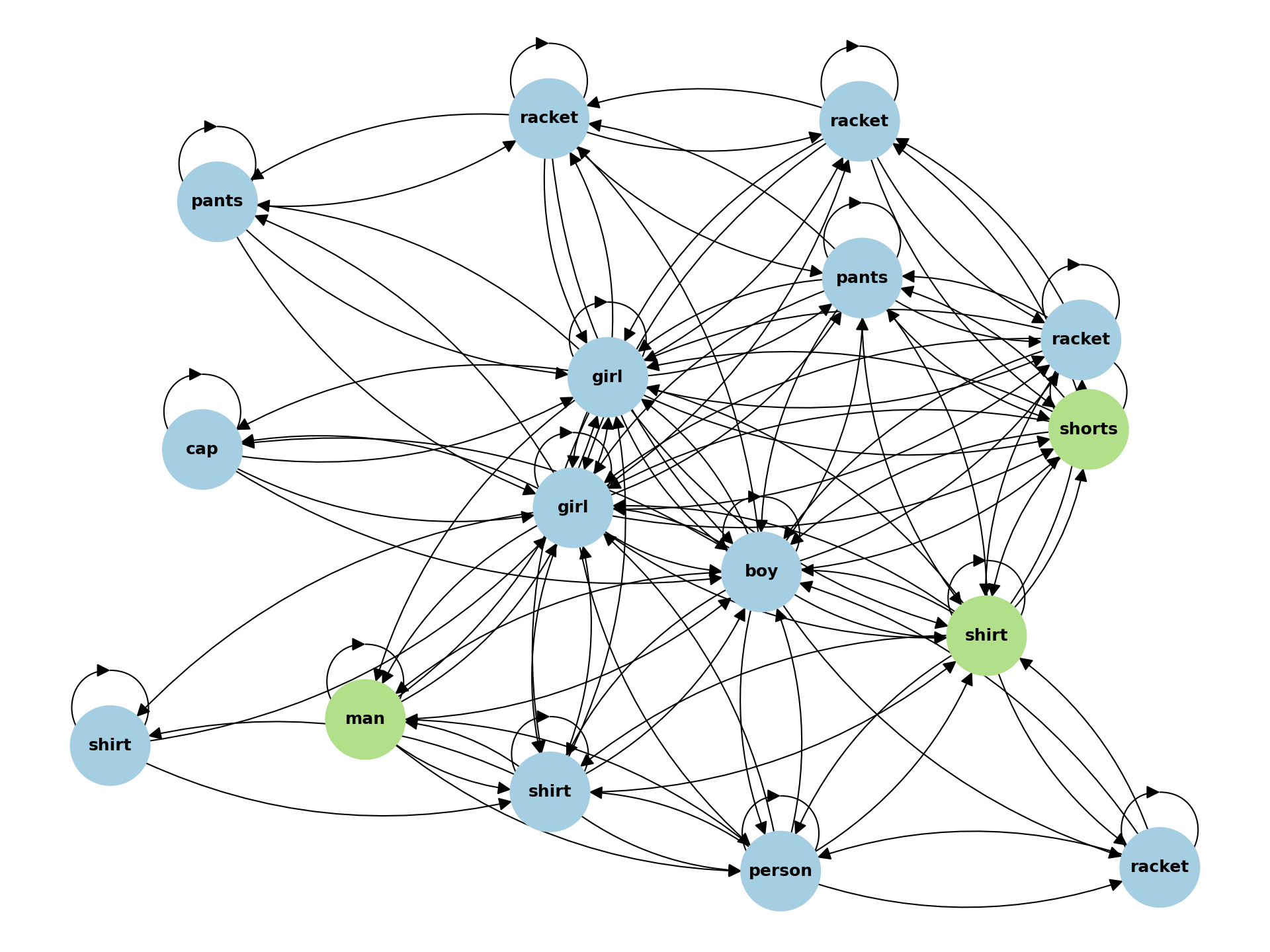}
        \caption{GNNExplainer.}
    \end{subfigure}
    \begin{subfigure}{\figsize\linewidth}
        \includegraphics[width=\textwidth, trim=0.4cm 0.4cm 0.4cm 0.4cm,clip]{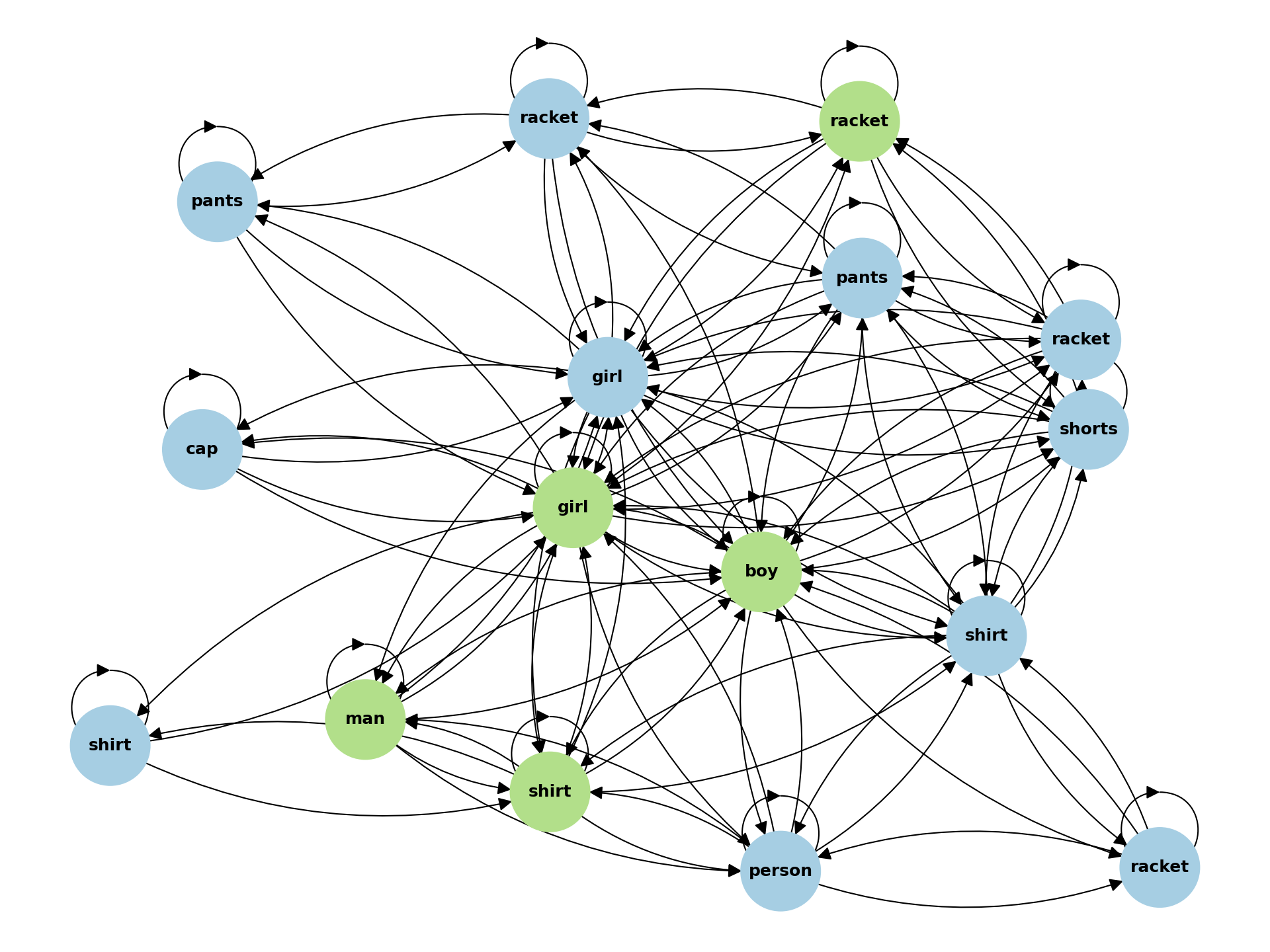}
        \caption{Random.}
    \end{subfigure}
    \begin{subfigure}{\figsize\linewidth}
        \includegraphics[width=\textwidth, trim=0.4cm 0.4cm 0.4cm 0.4cm,clip]{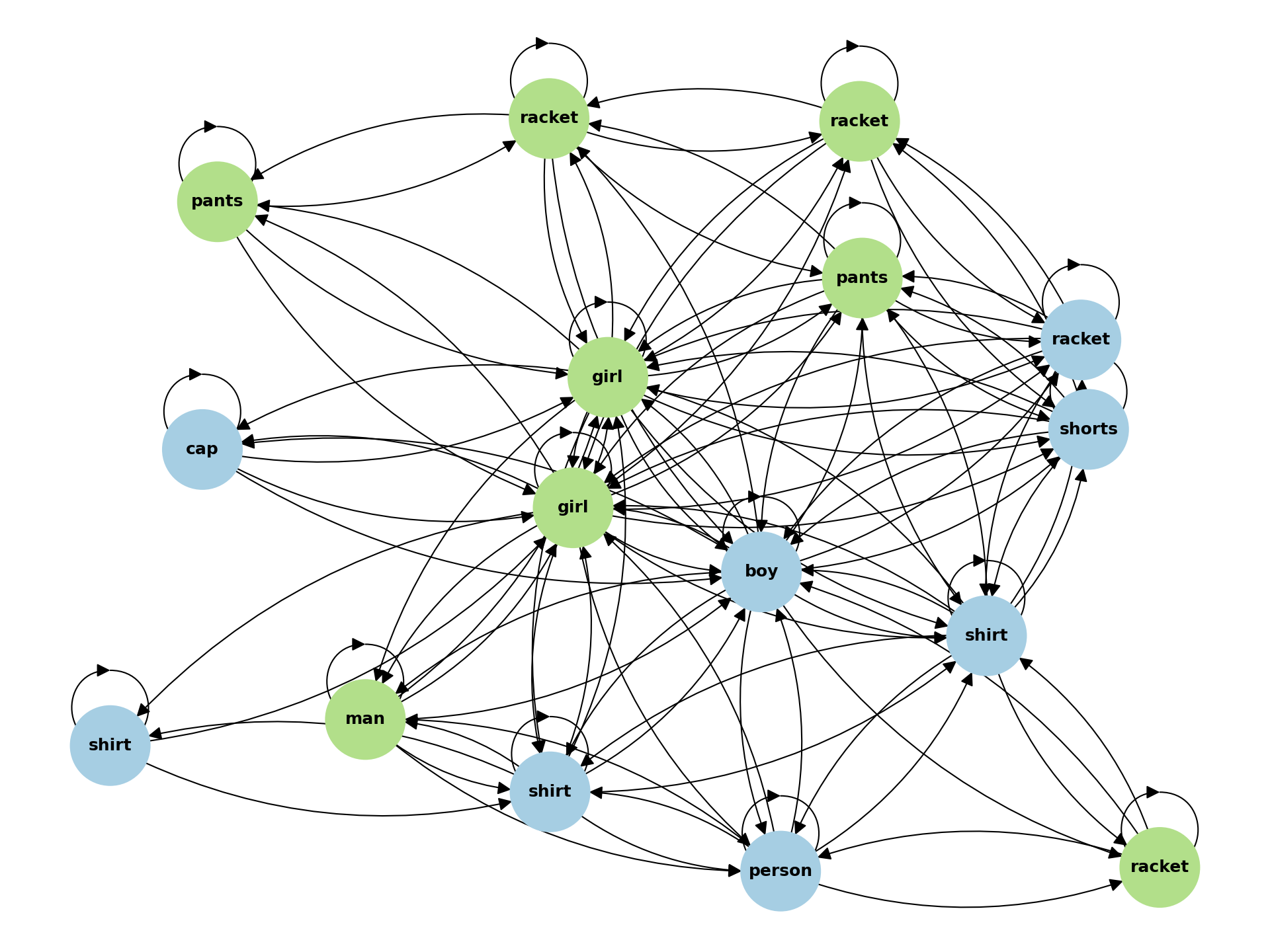}
        \caption{Integrated-Gradients.}
    \end{subfigure}
    \caption{Graph and image for question Id: 11389703. Question: \emph{Does the girl to the right of the person wear shorts?} Prediction: \emph{yes}. Ground-truth answer: \emph{yes}. Semantic type: \emph{relation}. Structural type: \emph{verify}. }
    \label{fig:example_5}
\end{figure*}

\begin{figure*}[htb]
    \centering
    \begin{subfigure}{\figsize\linewidth}
        \includegraphics[width=\textwidth, trim=0.4cm 0.4cm 0.4cm 0.4cm,clip]{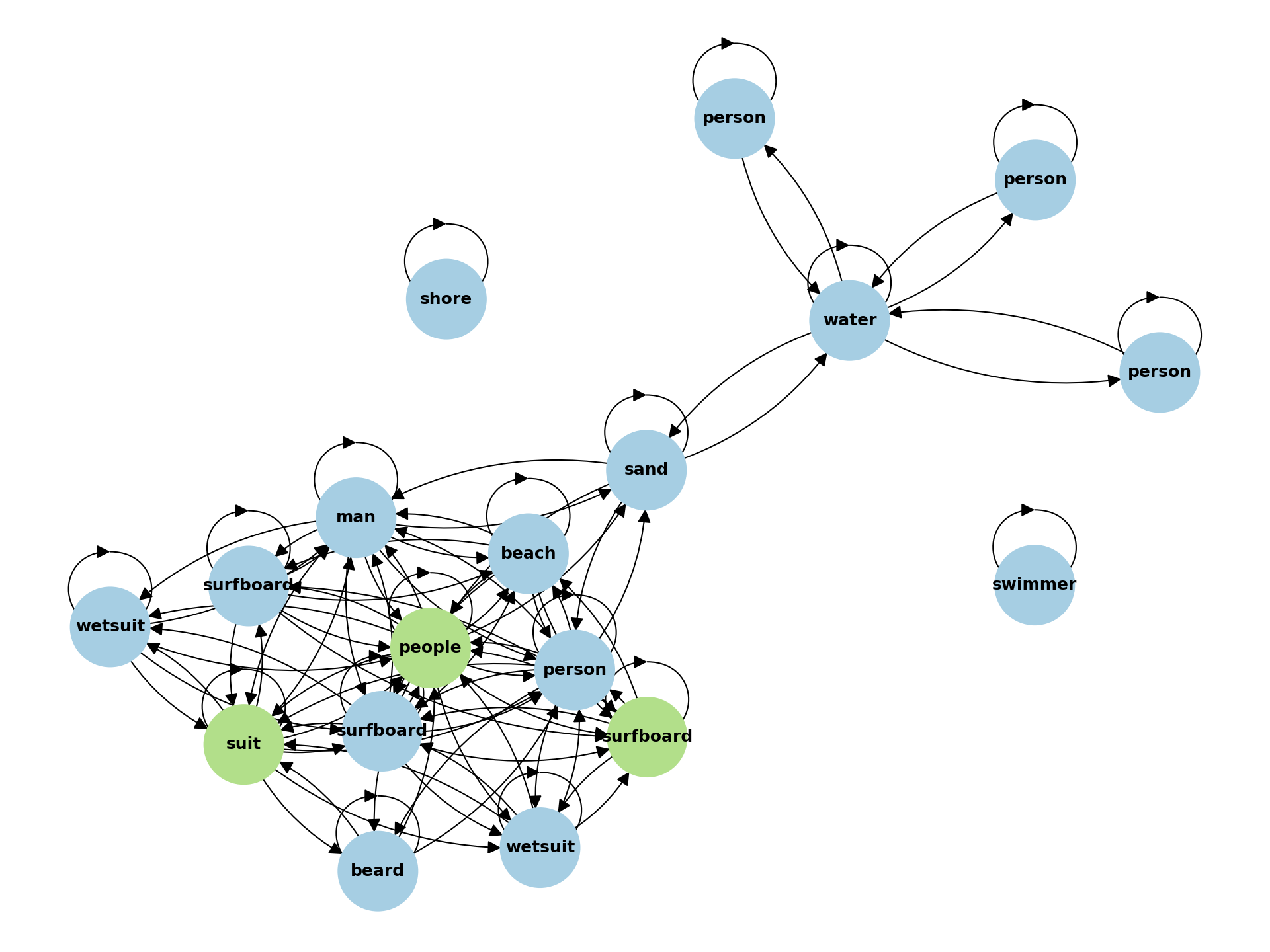}
        \caption{I-MLE.}
    \end{subfigure}
    \begin{subfigure}{\figsize\linewidth}
        \includegraphics[width=\textwidth, trim=0.4cm 0.4cm 0.4cm 0.4cm,clip]{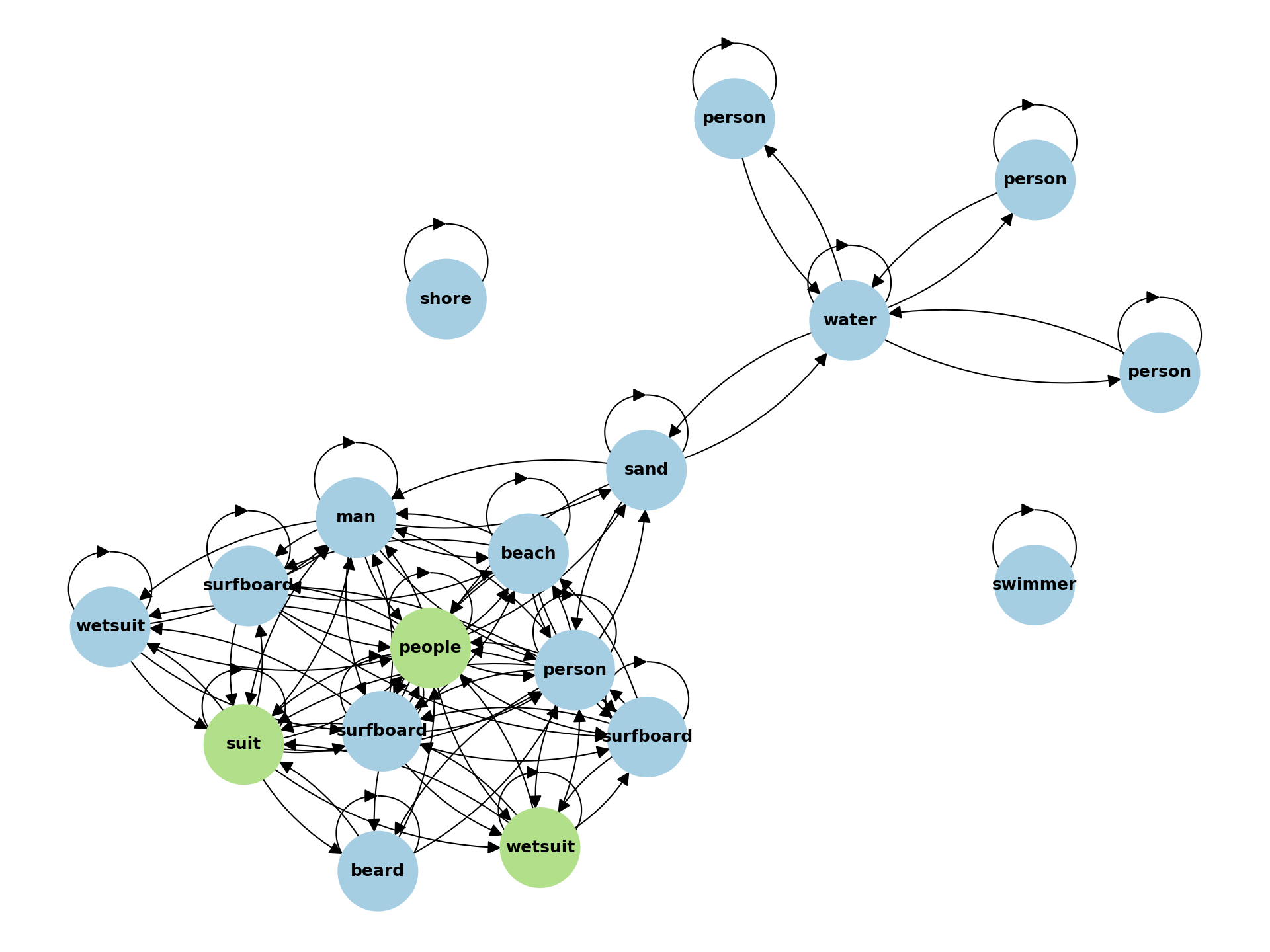}
        \caption{GNNExplainer.}
    \end{subfigure}
    \begin{subfigure}{\figsize\linewidth}
        \includegraphics[width=\textwidth, trim=0.4cm 0.4cm 0.4cm 0.4cm,clip]{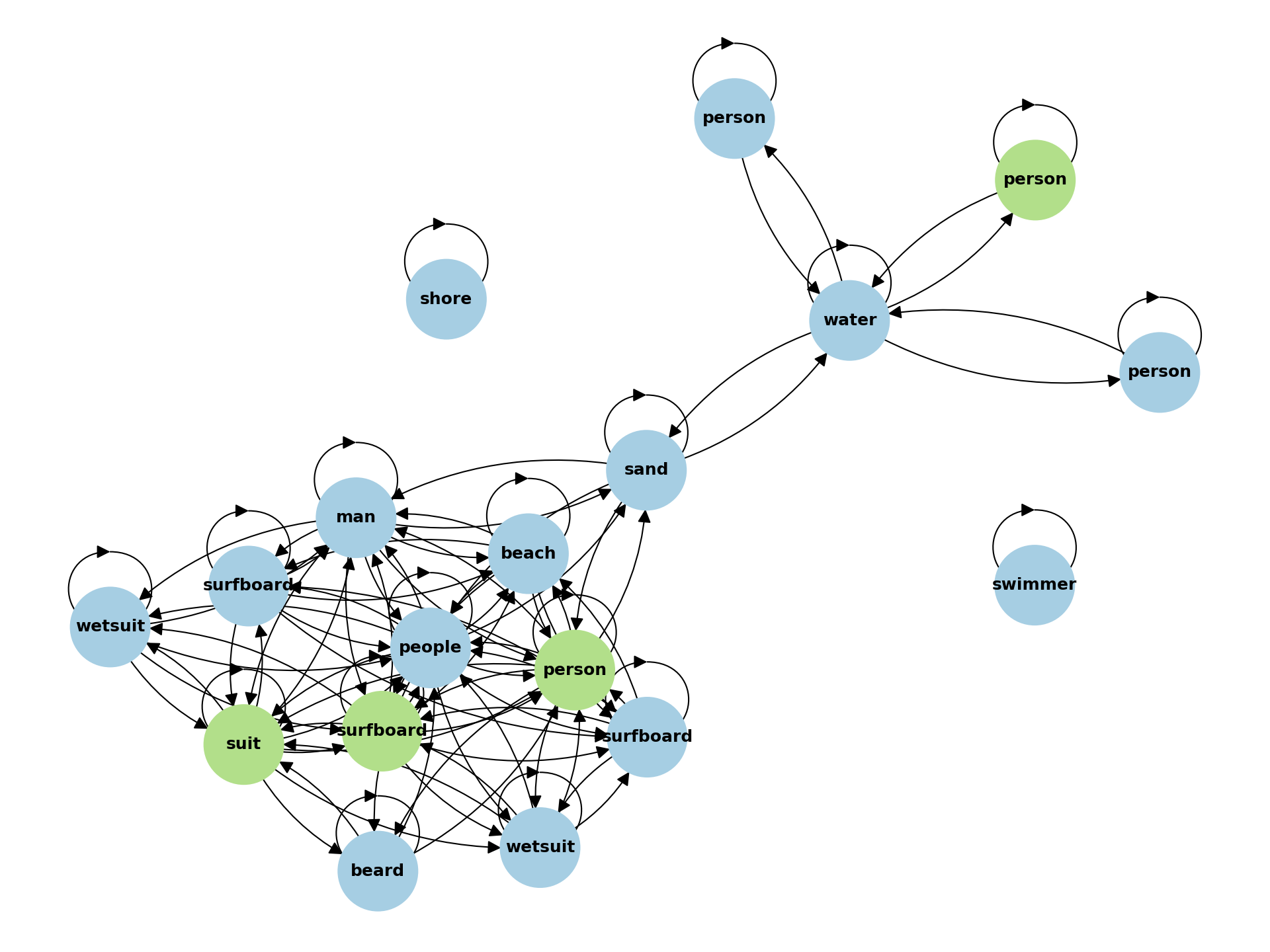}
        \caption{Random.}
    \end{subfigure}
    \begin{subfigure}{\figsize\linewidth}
        \includegraphics[width=\textwidth, trim=0.4cm 0.4cm 0.4cm 0.4cm,clip]{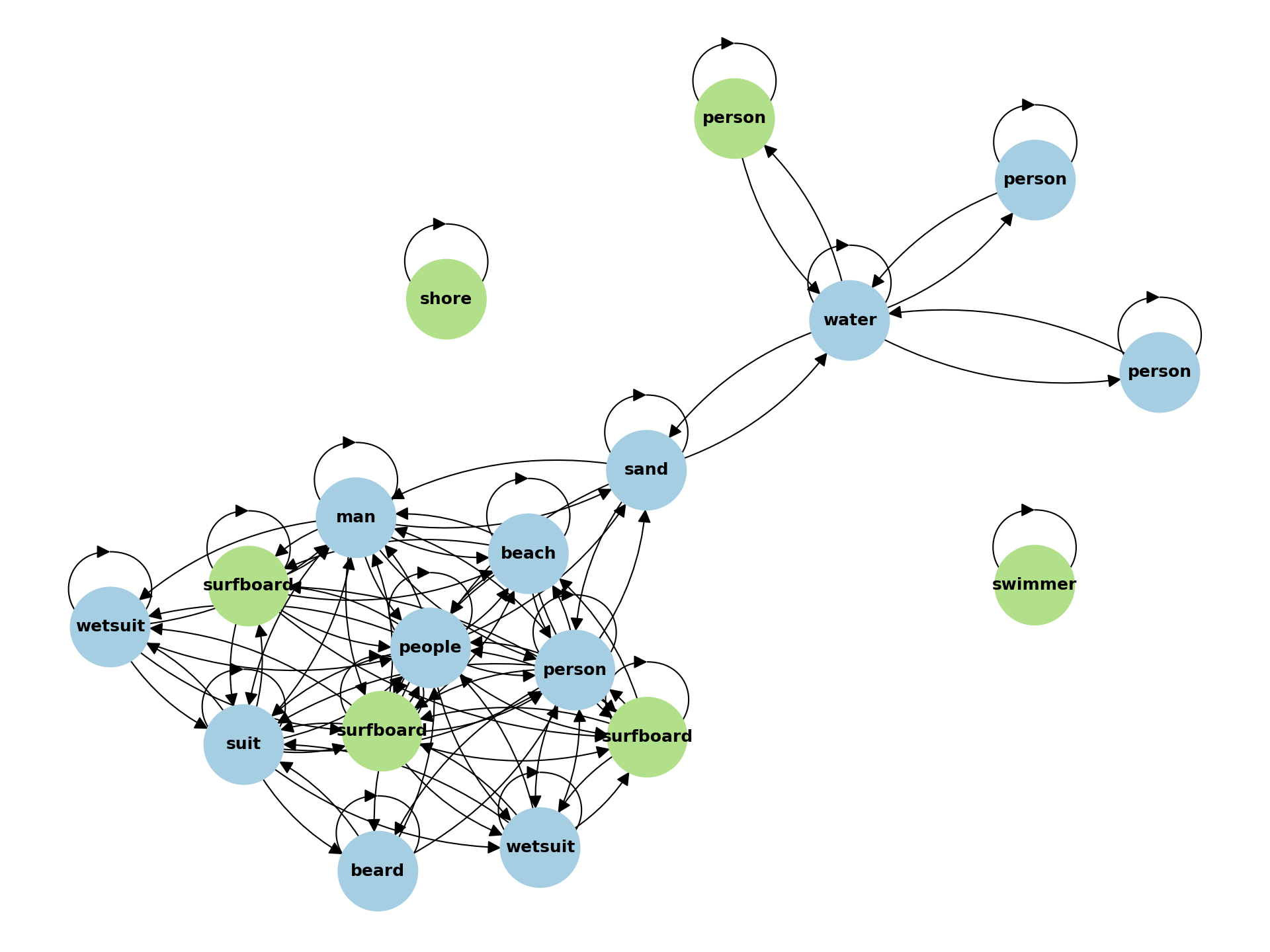}
        \caption{Integrated-Gradients.}
    \end{subfigure}
    \caption{Graph and image for question Id: 1662748. Question: \emph{Who is wearing a suit?} Prediction: \emph{people}. Ground-truth answer: \emph{people}. Semantic type: \emph{relation}. Structural type: \emph{query}.}
    \label{fig:example_8}
\end{figure*}

\twocolumn

\end{document}